\documentclass[twoside,11pt]{article}

\usepackage{blindtext}
\usepackage[abbrvbib, preprint]{jmlr2e}
\usepackage{jmlr2e}
\usepackage{hyperref}
\usepackage{url}
\usepackage{bm}
\usepackage{enumitem}  
\usepackage{graphicx} 
\usepackage{xcolor}
\usepackage{soul}
\usepackage{subcaption}
\usepackage{tabularx,ragged2e}
\usepackage{booktabs}
 \usepackage{multirow} 
\usepackage{amsmath}

\usepackage{lastpage}
\jmlrheading{23}{2022}{1-\pageref{LastPage}}{1/21; Revised 5/22}{9/22}{21-0000}{Author One and Author Two}

\ShortHeadings{Probabilistic Attribution For Large Language Models}{One and Two}
\firstpageno{1}

\begin{document}

\title{Probabilistic Attribution For Large Language Models}

\author{\name Shilpika \email shilpika@anl.gov \\
       \addr Argonne Leadership Computing Facility\\
       Argonne National Laboratory\\
       Lemont, IL, USA
       \AND
       \name Carlo Graziani \email cgraziani@anl.gov \\
       \addr Mathematics and Computer Science Division\\
       Argonne National Laboratory\\
       Lemont, IL, USA
       \AND
       \name Bethany Lusch \email blusch@anl.gov \\
       \addr Argonne Leadership Computing Facility\\
       Argonne National Laboratory\\
       Lemont, IL, USA
       \AND
       \name Venkatram Vishwanath \email venkat@anl.gov \\
       \addr Argonne Leadership Computing Facility\\
       Argonne National Laboratory\\
       Lemont, IL, USA
       \AND
       \name Michael E. Papka \email papka@anl.gov \\
       \addr Argonne Leadership Computing Facility\\
       Argonne National Laboratory\\
       Lemont, IL, USA \\
       \addr Department of Computer Science \\
       University of Illinois Chicago \\
       Chicago, IL, USA}

\editor{Pradeep Ravikumar, Carnegie Mellon University, 
Tong Zhang, University of Illinois Urbana-Champaign.}

\maketitle

\begin{abstract}
The generative nature of Large Language Models (LLMs) is reflected in the conditional probabilities they compute to sample each response token given the previous tokens. These probabilities encode the distributional structure that the model learns in training and exploits in inference. In this work, we use these probabilities to situate LLMs within the mathematical theory of stochastic processes. We use this framework to design a model-agnostic probabilistic token attribution measure, using Bayes rule to invert the next-token log-probabilities so as to capture the model's internal representation of the distribution over token sequences. The representation is independent of the model's computational structure. This representation yields the conditional probability of the response given the prompt, and of the response given the prompt with a token marginalized away. Our attribution score is the log of the ratio of these probabilities. We further compute the entropies of a single prompt’s token distributions, conditioned on the remaining context. The interplay between entropy and attribution score sheds light on LLM behavior. We evaluate 8 models across 7 prompts and investigate anomalies, token sensitivity, response stability, model stability, and training convergence, thereby improving interpretability and guiding users to focus on uncertain or unstable parts of the generation.
\end{abstract}


\section{Introduction}

The growing impact of AI on scientific modeling mandates a newly-rigorous approach to analyzing what AI models do, how they do it, and what uncertainties should be associated with AI-based predictions and decisions. Key concerns include appropriate model sizing (which directly impacts the energy cost of training), output validity/correctness, and output uncertainty, among others. These present substantial challenges, because the complexity of the models precludes the kind of direct interpretability that attends first-principles mathematical/statistical models, making it necessary to address such questions indirectly.

To enable the understanding of the decision-making process of AI models, particularly large language models, several techniques, frameworks, and methodologies have been developed, and these are referred to as Explainable Artificial Intelligence (XAI). XAI provides transparency into how inputs to ``black box" AI model generate a response. The motivation for XAI arises from the need to foster trust, regulation, accountability, fairness, and safety in AI applications, especially in high-stakes domains like healthcare, finance, and law, where opaque decision-making can have serious ethical and social consequences~\citep{916746, Rudin2019-od}.

XAI encompasses methods that can be broadly categorized into model-specific approaches, which are tailored to certain classes of models (for example, saliency maps and gradient-based visualizations in deep neural networks), and model-agnostic approaches, which can be applied universally regardless of the underlying algorithm~\citep{10.1016/j.inffus.2019.12.012}. Model-agnostic methods such as SHAP (SHapley Additive exPlanations) and LIME (Local Interpretable Model-agnostic Explanations) rely on input–output behavior rather than internal architecture, making them particularly versatile across different machine learning models~\citep{lundberg2017shap,ribeiro2016lime}. 

In this work, we situate the inference operations of large language models (LLMs) within the mathematical theory of stochastic processes. Using this framework, we develop a probabilistic attribution score (\texttt{AS}) for large-language models that is model-agnostic. The \texttt{AS} determines the importance of tokens in a prompt given a generated response. LLMs are fundamentally autoregressive, meaning they generate text one token at a time by sampling a computed probability distribution for the next output token conditioned on all the previous ones. LLMs work because they successfully infer statistical properties of token sequences from training corpuses and exploit those properties to make near-optimal prompt-response decisions. This process is directly grounded in the chain rule of probability, which allows the joint probability of a sequence of random variables to be factorized into a product of conditional probabilities. Stochastic process theory provides a useful mathematical framework for understanding those operations in a model-agnostic fashion that permits easy comparison between different types of LLMs. This framework also permits useful probabilistic analyses of LLM responses, of which \texttt{AS} is an example.\\

Our work has the following contributions:
\begin{itemize}
    \item[\bf C1.]{ \texttt{AS} is model-agnostic and is applied to text-generation tasks, although it can be extended to any NLP-based task.}
    \item[\bf C2.]{Our method is parameter-free, which makes it less sensitive to arbitrary user choices and improves reproducibility and objectivity, making it suitable for hyperparameter-tuning of other XAI methods.}
    \item[\bf C3.]{Our contextual and replacement entropy are measures of model stability, and can indicate how well these models converged to their training data distribution.}
    \item[\bf C4.]{\texttt{AS} has built-in sensitivity, as it is based on statistical first principles and thus evaluates the correctness of the explainable results as part of its core methodology.}
\end{itemize}

 The remainder of this paper is organized as follows. The Background section reviews the key concepts, prior work, and theoretical foundations relevant to the study. The LLMs as Stochastic Processes, Experiments, and Evaluation sections then describe the proposed approach, experimental setup, implementation details, comparative analysis,  and the criteria used to assess performance. Finally, the Conclusion summarizes the main findings, discusses their implications, and outlines potential directions for future research. The code is available at: \textit{https://github.com/sshilpika/probabilistic-attribution-score/}.

\section{Background: Explainable AI for NLP and Large Language Models}

Explainable artificial intelligence (XAI) in natural language processing (NLP) has evolved from post hoc interpretability methods for linear models and early neural architectures into a diverse methodological landscape tailored to Transformer encoders and contemporary large language models \citep{luo2024xaiLLM,rogers2020bertology}. Current research spans three broad levels: local explanations of individual predictions, global analyses of model behavior and internalized knowledge, and mechanistic interpretability aimed at identifying the internal computational structures that give rise to model outputs \citep{luo2024xaiLLM,elhage2021transformerCircuits}. 

Early work in NLP XAI focused on local feature-attribution methods designed to estimate which input tokens or n-grams most influenced a model's prediction. LIME represented a major advance by explaining predictions through locally fitted interpretable surrogate models, often using perturbation-based token deletion \citep{ribeiro2016lime}. Closely related Shapley-value-based methods, particularly SHAP, provided a more principled additive framework for attribution and became widely adopted in text-based interpretability pipelines \citep{lundberg2017shap}. For neural NLP models, gradient-based methods such as Integrated Gradients became especially influential due to their computational scalability and axiomatic grounding \citep{sundararajan2017ig}. Toolkits such as Captum further standardized the implementation and comparison of attribution methods in PyTorch-based workflows \citep{kokhlikyan2020captum}. However, it soon became clear that token-level attributions could be unstable under paraphrase, and sensitive to perturbation strategy \citep{DBLP:journals/corr/abs-2108-04990}. 

These limitations motivated the development of rationale-based explanations, which identify spans of text intended to justify a prediction. This line of work was consolidated by benchmarks such as ERASER, which distinguished plausibility, or agreement with human-annotated rationales, from faithfulness, or the extent to which a rationale genuinely supports the model’s prediction \citep{deyoung2020eraser}. This distinction has become foundational in LLM explainability, where persuasive explanations may still fail to reflect the actual computational basis of a response \citep{luo2024xaiLLM}. 

Attention-based interpretability introduced another major debate. Because attention maps are visually intuitive, they were initially treated as natural explanations of Transformer behavior. However, Jain and Wallace showed that attention weights often correlate weakly with other importance measures and that substantially different attention distributions can yield similar predictions \citep{jain2019attention}. Subsequent work argued that the explanatory value of attention depends on the definition of explanation and the evaluation protocol employed \citep{wiegreffe2019attention}.

In parallel, global interpretability research examined what linguistic knowledge models encode, where it is stored, and how it evolves across layers. This tradition, often referred to as “BERTology,” employed probing classifiers, representational analyses, and behavioral tests to characterize syntactic and semantic information in model representations \citep{rogers2020bertology}. More recently, mechanistic interpretability has extended this agenda by attempting to reverse-engineer Transformers into circuits, features, and information pathways \citep{elhage2021transformerCircuits}. 

In the LLM era, explainability must also address prompt sensitivity, self-generated rationales, and retrieval provenance in retrieval-augmented systems \citep{luo2024xaiLLM}. Overall, the field can be organized along several dimensions: intrinsic versus post hoc methods, local versus global scope, the object of explanation, and the criteria of evaluation. Across these paradigms, the central challenge remains unchanged: producing explanations that are not merely plausible, but causally faithful, stable, and useful in real-world deployment \citep{luo2024xaiLLM,deyoung2020eraser}. Our work is model-agnostic and does not attempt to ascribe anything relating to ``knowledge representation" or ``model decision process" because it only concerns itself with text-distributional properties of LLM inference. 
\section{LLMs as Stochastic Processes}\label{sec:stochastic_processes}

Our approach to LLMs is probabilistic, and explicitly designed to draw a veil
over the computational transformer model details. Instead, we focus here on what
the transformer architecture explicitly purports to model: conditional
probabilities of tokens, conditioned on the tokens that preceded them in the
token sequence. In effect, a trained LLM defines a probability distribution over texts, which are sequences of tokens.

Viewed in this light, one can recognize a connection between LLMs and a venerable subject in mathematical statistics: the theory of stochastic processes. This connection is interesting, because it offers the possibility of situating LLMs within a well-developed body of mathematical theory, and of exploiting that theory to gain insight into LLMs and their outputs.

We remind readers of the definition of such objects. A \emph{stochastic process} is defined as a collection of random variables $[X_t:t\in T]$ on a probability space $(\Omega,\mathcal{F},\mathrm{Pr})$, where $T$ is an arbitrary index set \cite[][p. 482]{billingsley1995probability}. Recall further the elements of a probability space: $\Omega$ is a set of outcomes, called the \emph{sample space}, $\mathcal{F}$ is a $\sigma$-field of sets of elements of $\Omega$ (the measurable sets), and $\mathrm{Pr}:\mathcal{F}\mapsto [0,1]$ is a probability measure satisfying $\mathrm{Pr}(\Omega)=1$, $\mathrm{Pr}(\emptyset)=0$, and $\mathrm{Pr}\left(\cup_{k}A_k\right)=\sum_k P(A_k)$ for any countable collection of disjoint sets $A_k\in\mathcal{F}$.

In the context of LLMs, the index set is the natural numbers including zero, $T=\mathbb{N}_0$. The random variables correspond to the tokens issued during LLM inference at each location of a text sequence.  To further press the correspondence between the operations of an LLM and stochastic process theory, it is necessary to introduce some notation.

\subsection{Notation}

The objects of interest here are sequences of language tokens drawn from the
distribution over texts learned by an LLM during training. We denote position in
a token sequence by a Greek index such as $\mu$, distinguishing such indices
from Roman indices such as $l$ that denote the identity of a token. The size of
the vocabulary is fixed and denoted by $V$, so that a token ID (TID) index $l$
satisfies $0\le l<V$.

We will denote random variables by upper-case letters, and their realizations
by lower-case letters. So, in particular, the symbol $L_{\mu}$ represents
a random variable describing the distribution over possible TID values
of the token occurring at position $\mu$ in a sequence, while the
symbol $l_{\mu}$ ($0\le l_{\mu}<V$) represents a realization of
$L_{\mu}$. 

A finite-length token sequence will be represented
by a boldface letter. So, for example, we may write random variables such as
$\bm{L}=\left[L_{0},L_{1},\ldots,L_{M-1}\right]$ to represent sequences of random variables of length $M$. Similarly, we write
$\bm{l}=\left[l_{0},l_{1},\ldots,l_{M-1}\right]$ for a realized sequence of TIDs
of length $M$. Note that the elements of such finite-length sequences need not be contiguous in sequential index $\mu$, so that, for example $\bm{L}=\left[L_5,L_9,L_{22}\right]$ is a valid finite-length sequence of random variables, and may be realized by corresponding assignments $\bm{l}=\left[l_5,l_9,l_{22}\right]$. We denote the length of a finite-length sequence $\bm{L}$ by $|\bm{L}|$.

We may now further amplify the correspondence between the operations of LLMs and the above definition of a stochastic process.
The infinite-length sequence of random variables  $\mathcal{L}\equiv[L_\mu:\mu\in\mathcal{N}_0]=[L_0,L_1,\ldots]$ corresponds to the collection $[X_t:t\in T]$ in the definition of a stochastic process. The sample space $\Omega$ in the definition corresponds to the set of all possible infinite sequences $\omega=(l_0,l_1,l_2,\ldots)$ of TIDs.

As usual in probability theory, an ``event'' is a set belonging to the
$\sigma$-field of measurable sets $\mathcal{F}$, which by the definition of a
$\sigma$-field is closed under set complementation and countable set unions. We
construct $\mathcal{F}$ as follows: A set expression such as $\left\{
\bm{L}=\bm{l}\right\} $ describes the event ``The realization of the finite-length sequence
$\bm{L}$ is $\bm{l}$,'' and corresponds to the set of all sequences
$\mathcal{L}$ such that the corresponding TID assignments hold, while all other
elements of $\mathcal{L}$ are unspecified (i.e. unobserved). Denote by $F$ the collection of
all such sets, their finite unions, $\Omega$, and the empty set $\emptyset$.  $F$ is closed under complementation and finite
unions, and is therefore a field.  We define $\mathcal{F}$ as the
$\sigma$-field generated by $F$, $\mathcal{F}=\sigma(F)$. It is, by definition,
the smallest of all $\sigma$-fields containing $F$ \cite[][pp.
21-22]{billingsley1995probability}.

With this standard set notation for events, the event describing ``event $A$
\emph{and} event $B$'' is succinctly written $A\cap B$, whereas
``event $A$ \emph{or} event $B$'' is $A\cup B$.

In what follows, it will be convenient to have a notation for certain types of subsequences. With
the length-$M$ sequence $\bm{L}=\left[L_{0},L_{1},\ldots,L_{M-1}\right]$, we will write $\bm{L}_{<\mu}\equiv\left[L_{0},\ldots,L_{\mu-1}\right]$,
$\bm{L}_{\le\mu}\equiv\left[L_{0},\ldots,L_{\mu}\right]$, $\bm{L}_{>\mu}\equiv\left[L_{\mu+1},\ldots,L_{M-1}\right]$,
and similarly for the sequence realizations $\bm{l}_{<\mu}$, $\bm{l}_{\le\mu}$,
$\bm{l}_{>\mu}$. The notation $\bm{L}-L_{\mu}\equiv\text{\ensuremath{\bm{L}_{<\mu}}}\cap\bm{L}_{>\mu}$
then represents the finite subsequence
over the $L_{\nu}$, $0\le\nu<M$, $\nu\neq\mu$. This helps us represent
events such as $\left\{ \bm{L}-L_{\mu}=\bm{l}-l_{\mu}\right\} $,
the set of sequences wherein all tokens
in positions $\nu<M$ except $l_{\mu}$ have realized values, while
the values of $l_{\mu}$ and of $l_{\nu}$, $\nu\ge M$ are undetermined.

\subsection{The Probability Measure}

The striking thing about the view of LLMs as stochastic processes is that the probability measure $\mathrm{Pr}$ that enters the probability space $(\Omega,\mathcal{F},\mathrm{Pr})$ is provided by the LLM itself, constructed from its training data.

LLMs operate sequentially, yielding vectors of categorical probabilities
for the next token in a sequence conditioned on all previous tokens.
That is, they supply
\begin{equation}
\mathrm{Pr}\left(\left\{L_{\mu}=l_\mu\right\}|\left\{\bm{L}_{<\mu}=\bm{l}_{<\mu}\right\}\right)\label{eq:Sequential}
\end{equation}
where the $L_{\mu}$ are the tokens that make up a sequence. By the
chain rule of probability one can directly form response sequence
probabilities via
\begin{eqnarray}
\mathrm{Pr}\left(\left\{\bm{L}_{\ge\mu}=\bm{l}_{\ge\mu}\right\}|\left\{\bm{L}_{<\mu}=\bm{l}_{<\mu}\right\}\right) & = & 
\mathrm{Pr}\left(\left\{L_{\mu}=l_\mu\right\}|\left\{\bm{L}_{<\mu}=\bm{l}_{<\mu}\right\}\right)\nonumber\\
& &\times\mathrm{Pr}\left(\left\{L_{\mu+1}=l_{\mu+1}\right\}|\left\{\bm{L}_{<\mu+1}=\bm{l}_{<\mu+1}\right\}\right)\times\ldots\nonumber \\
 &  & \times\mathrm{Pr}\left(\left\{L_{|\bm{L}|-1}=l_{|\bm{L}|-1}\right\}|\left\{\bm{L}_{<|\bm{L}|-1}=\bm{l}_{<|\bm{L}|-1}\right\}\right).\label{eq:Sequential_2}
\end{eqnarray}
These conditionals are used to sample (according to some sampling strategy) tokens sequentially, so as to produce extended responses $\bm{l}_{\ge\mu}$ to prompts $\bm{l}_{<\mu}$.

When an LLM inference process is run locally (i.e. not through a chatbot or a vendor API) these probabilities are easily available as log (unnormalized) probabilities, which can be converted to normalized probability vectors by a softmax operation. As a consequence we gain access to the LLM's approximation to the distribution over its training texts, which we may reconstruct according to Equation~(\ref{eq:Sequential_2}).

As we demonstrate in the next section, once we have this reconstruction we are no longer constrained to conditional expressions such as those in Equations (\ref{eq:Sequential}--\ref{eq:Sequential_2}), wherein all sequences are complete and contiguous, and the sequences on the left of the conditional bar follow immediately those on the right.  In fact it is possible to construct \emph{arbitrary} expressions $Pr(\bm{S}_1|\bm{S}_2)$ where $\bm{S}_1$ and $\bm{S}_2$ are non-contiguous, non-sequential sets of tokens, simply by appropriately prompting an LLM i.e. by coercing it to sample the required tokens at the specified sequence locations, and siphoning off the emitted vectors of log-probability. We now illustrate this statement by deriving an expression for a score that measures the sensitivity of an LLM response to individual tokens in a prompt.

\subsection{Probabilistic Attribution Score}

This section describes our approach for attributing importance to prompt
tokens in light of the LLM response to the prompt.

The basic idea is that if we have a prompt $\bm{p}$ that induces
a response $\bm{r}$ sampled from the distribution $\mathrm{Pr}\left(\left\{\bm{R}=\bm{r}\right\}|\left\{\bm{P}=\bm{p}\right\}\right)$
(where $\bm{p}$ and $\bm{r}$ are contiguous token sequences), then we would
like a measure of attribution of the importance to the $\mu$-th prompt
token $p_{\mu}$ to $\bm{r}$ given by
\begin{equation}
A_{\mu}\equiv\log\frac{\mathrm{Pr}\left(\left\{\bm{R}=\bm{r}\right\}|\left\{\bm{P}=\bm{p}\right\}\right)}
{\mathrm{Pr}\left(\left\{\bm{R}=\bm{r}\right\}|\left\{\bm{P}-P_{\mu}=\bm{p}-p_\mu\right\}\right)},\label{eq:Attribution}
\end{equation}
where, again, the notation ``$\left\{\bm{P}-P_{\mu}=\bm{p}-p_\mu\right\}$'' means ``the token sequence
$\bm{p}$ omitting the value of token $p_{\mu}$ occurring at position
$\mu$.'' Note that in $\bm{p}-p_{\mu}$, $p_{\mu}$ itself is not
dropped---that is, the sequence $\bm{p}$ is not shortened by 1---but
$p_{\mu}$ could have attained any value in the token vocabulary,
instead of being assigned the value given in $\bm{p}$. In effect, the token $p_\mu$ is masked, in the sense that the information that it contributes to the conditional probability for $\left\{\bm{R}=\bm{r}\right\}$ is dropped.

A large value of $A_{\mu}$ could indicate high importance of $p_{\mu}$
to $\bm{r}$, since it would follow that the inclusion of $p_{\mu}$
dramatically improves the probability of observing $\bm{r}$. More to the point,
the distribution of $A_{\mu}$ over $\mu$ yields an idea of which
tokens in the prompt were the chief perpetrators of the response.

There is a difficulty in computing the expression in Equation (\ref{eq:Attribution}),
which is that LLMs do not directly output expressions
such as $\mathrm{Pr}\left(\left\{\bm{R}=\bm{r}\right\}|\left\{\bm{P}-P_{\mu}=\bm{p}-p_\mu\right\}\right)$. Instead,
they operate sequentially, yielding vectors of categorical probabilities
for the next token in a sequence conditioned on all previous tokens.
That is, they supply expressions such as Equation (\ref{eq:Sequential}) or (\ref{eq:Sequential_2}).

The problem, then, is to construct expressions for the denominator
of Equation (\ref{eq:Attribution}) using nothing but expressions
such as Equation (\ref{eq:Sequential}) or (\ref{eq:Sequential_2}).
It turns out that this is indeed possible [\textbf{C1, C2, C4}].

Per Equation (\ref{eq:Attribution}), we require an expression for
the denominator $D$,
\begin{eqnarray*}
D&\equiv&\mathrm{Pr}\left(\left\{ \bm{R}=\bm{r}\right\} |\left\{ \bm{P}-P_{\mu}=\bm{p}-p_{\mu}\right\} \right)\\
&=&\mathrm{Pr}\left(\left\{ \bm{R}=\bm{r}\right\} |\left\{ \bm{P}_{<\mu}=\bm{p}_{<\mu}\right\} \cap\left\{ \bm{P}_{>\mu}=\bm{p}_{>\mu}\right\} \right).
\end{eqnarray*}
The required expressions must only contain factors such
as those in Equations (\ref{eq:Sequential}--\ref{eq:Sequential_2})
so that they may be read out of the LLM on a side-channel that reports
categorical probabilities over the next token in the sequence. Assume that the response follows the prompt sequentially, with no
gap between them, and that both prompt and response are composed of contiguous sequence locations. We start with

\begin{eqnarray*}
D & = & \sum_{p_{\mu}^{\prime}=0}^{V-1}\mathrm{Pr}\left(\left\{ \bm{R}=\bm{r}\right\} \cap\left\{ P_{\mu}=p_{\mu}^{\prime}\right\} |\left\{ \bm{P}_{<\mu}=\bm{p}_{<\mu}\right\} \cap\left\{ \bm{P}_{>\mu}=\bm{p}_{>\mu}\right\} \right),
\end{eqnarray*}
since this expresses the view that $P_{\mu}$ may realize any value
in the vocabulary. Now, define 
\[
\bm{p}^{\prime}\equiv\left[p_{0},\ldots,p_{\mu-1},p_{\mu}^{\prime},p_{\mu+1},\ldots,p_{M-1}\right].
\]
Since 
\[
\left\{ \bm{P}_{<\mu}=\bm{p}_{<\mu}\right\} \cap\left\{ \bm{P}_{>\mu}=\bm{p}_{>\mu}\right\} \cap\left\{ P_{\mu}=p_{\mu}^{\prime}\right\} =\left\{ \bm{P}=\bm{p}^{\prime}\right\} ,
\]
we may write
\begin{eqnarray}
D & = & \sum_{p_{\mu}^{\prime}=0}^{V-1}\mathrm{Pr}\left(\left\{ \bm{R}=\bm{r}\right\} |\left\{ \bm{P}=\bm{p}^{\prime}\right\} \right)\nonumber \\
 &  & \times\mathrm{Pr}\left(\left\{ P_{\mu}=p_{\mu}^{\prime}\right\} |\left\{ \bm{P}_{<\mu}=\bm{p}_{<\mu}\right\} \cap\left\{ \bm{P}_{>\mu}=\bm{p}_{>\mu}\right\} \right).\label{eq:chain_1}
\end{eqnarray}
The first factor in the summand is in the required form. As to the
second factor, we may apply Bayes' rule
\begin{eqnarray}
B(p^\prime_\mu) & \equiv & \mathrm{Pr}\left(\left\{ P_{\mu}=p_{\mu}^{\prime}\right\} |\left\{ \bm{P}_{<\mu}=\bm{p}_{<\mu}\right\} \cap\left\{ \bm{P}_{>\mu}=\bm{p}_{>\mu}\right\} \right)\nonumber \\
 & = & \frac{\mathrm{Pr}\left(\left\{ \bm{P}_{>\mu}=\bm{p}_{>\mu}\right\} |\left\{ \bm{P}_{\le\mu}=\bm{p}^\prime_{\le\mu}\right\} \right)\times\mathrm{Pr}\left(\left\{ P_{\mu}=p_{\mu}^{\prime}\right\} |\left\{ \bm{P}_{<\mu}=\bm{p}_{<\mu}\right\} \right)}{\sum_{p_{\mu}^{\prime\prime}=0}^{V-1}\mathrm{Pr}\left(\left\{ \bm{P}_{>\mu}=\bm{p}_{>\mu}\right\} |\left\{ \bm{P}_{\le\mu}=\bm{p}^{\prime\prime}_{\le\mu}\right\} \right)\times\mathrm{Pr}\left(\left\{ P_{\mu}=p_{\mu}^{\prime\prime}\right\} |\left\{ \bm{P}_{<\mu}=\bm{p}_{<\mu}\right\} \right)},\label{eq:bayes}
\end{eqnarray}
wherein all factors appearing on the right are in the required form.
Combining Equations (\ref{eq:chain_1}) and (\ref{eq:bayes}),
\begin{eqnarray}
D & = & \sum_{p_{\mu}^{\prime}=0}^{V-1}\mathrm{Pr}\left(\left\{ \bm{R}=\bm{r}\right\} |\left\{ \bm{P}=\bm{p}^{\prime}\right\} \right)\times B(p^\prime_\mu)\nonumber \\
\label{eq:Denom}
\end{eqnarray}

It follows that our attribution measure for prompt location $\mu$
is
\begin{eqnarray}
A_{\mu} & = & \log \frac{\mathrm{Pr}\left(\left\{ \bm{R}=\bm{r}\right\} |\left\{ \bm{P}=\bm{p}\right\} \right)}{D}\nonumber \\
&=&\log\frac{X}{Y},
\label{eq:Attribution_2}
\end{eqnarray}
where
\begin{eqnarray}
X&\equiv&\mathrm{Pr}\left(\left\{ \bm{R}=\bm{r}\right\} |\left\{ \bm{P}=\bm{p}\right\} \right)
\times\sum_{p_{\mu}^{\prime}=0}^{V-1}\mathrm{Pr}\left(\left\{ \bm{P}_{>\mu}=\bm{p}_{>\mu}\right\} |\left\{ \bm{P}_{\le\mu}=\bm{p}^\prime_{\le\mu}\right\} \right)
\nonumber\\
&&\hphantom{\mathrm{Pr}\left(\left\{ \bm{R}=\bm{r}\right\} |\left\{ \bm{P}=\bm{p}\right\} \right)
\times\sum_{p_{\mu}^{\prime}=0}^{V-1}}
\times\mathrm{Pr}\left(\left\{ P_{\mu}=p_{\mu}^{\prime}\right\} |\left\{ \bm{P}_{<\mu}=\bm{p}_{<\mu}\right\} \right),\label{eq:Attribution_3}\\
Y&\equiv&\sum_{p_{\mu}^{\prime}=0}^{V-1}\mathrm{Pr}\left(\left\{ \bm{R}=\bm{r}\right\} |\left\{ \bm{P}=\bm{p}^{\prime}\right\} \right)\times\mathrm{Pr}\left(\left\{ \bm{P}_{>\mu}=\bm{p}_{>\mu}\right\} |\left\{ \bm{P}_{\le\mu}=\bm{p}^\prime_{\le\mu}\right\} \right)
\nonumber\\
&&\hphantom{\sum_{p_{\mu}^{\prime}=0}^{V-1}\mathrm{Pr}\left(\left\{ \bm{R}=\bm{r}\right\} |\left\{ \bm{P}=\bm{p}^{\prime}\right\} \right)\times}
\times\mathrm{Pr}\left(\left\{ P_{\mu}=p_{\mu}^{\prime}\right\} |\left\{ \bm{P}_{<\mu}=\bm{p}_{<\mu}\right\} \right)\label{eq:Attribution_4}.
\end{eqnarray}

\subsection{Contextual Entropies}

A second useful indicator of prompt token sensitivity can be obtained by considering the entropy of each prompt sequence location, conditioned on the remaining prompt tokens, or on the entire remaining context, including the response.

The entropy of a categorical distribution $\bm{q}$ with components $[\bm{q}]_l\equiv q_l\ge 0$, $l=0,\ldots,V-1$ satisfying $\sum_{l=0}^{V-1}q_l=1$ is given by the formula
\begin{equation}
S(\bm{q})=-\sum_{l=0}^{V-1}q_l\log q_l.\label{eq:entropy}
\end{equation}
Its significance is that it is an information-theoretic measure of the concentration of the distribution $\bm{q}$ over some set of outcomes. In the case of complete concentration on one outcome, $q_l=1$, $q_{m\neq l}=0$ we have $S(\bm{q})=0$, a lower bound, whereas in the case of maximum dispersion $\forall l:q_l=1/V$ we have $S(\bm{q})=\log V$, an upper bound \citep{CoverThomas}.

We set the probabilities $\bm{q}$ in Equation (\ref{eq:entropy}) to either the prompt-conditioned probabilities at sequence location $\mu$, i.e.
\begin{eqnarray}
\left[\bm{q}^{(P)}_\mu\right]_{p_\mu}&\equiv&
\mathrm{Pr}\left(
P_\mu=p_\mu|
\{\bm{P}_{<\mu}=\bm{p_{<\mu}}\}\cap\left\{\bm{P}_{>\mu}=\bm{p}_{>\mu}\right\}
\right)\nonumber\\
&=&
\frac{\mathrm{Pr}\left(\{\bm{P}_{\ge\mu}=\bm{p_{\ge\mu}}\}|
\left\{\bm{P}_{<\mu}=\bm{p}_{<\mu}\right\}\right)}
{\sum_{p^\prime_\mu=0}^{V-1}\mathrm{Pr}\left(
\left\{\bm{P}_{\ge\mu}=\bm{p_{\ge\mu}^\prime}\right\}|
\left\{\bm{P}_{<\mu}=\bm{p}_{<\mu}\right\}\right)},
\end{eqnarray}
or to the prompt+response-conditioned probabilities sequence location $\mu$, i.e.
\begin{eqnarray}
\left[\bm{q}^{(P+R)}_\mu\right]_{p_\mu}&\equiv&
\mathrm{Pr}\left(
P_\mu=p_\mu|
\{\bm{P}_{<\mu}=\bm{p_{<\mu}}\}\cap
\left\{\bm{P}_{>\mu}=\bm{p}_{>\mu}\right\}\cap
\left\{\bm{R}=\bm{r}\right\}
\right)\nonumber\\
&=&
\frac{\mathrm{Pr}\left(\{\bm{P}_{\ge\mu}=\bm{p_{\ge\mu}}\}\cap
\left\{\bm{R}=\bm{r}\right\}|
\left\{\bm{P}_{<\mu}=\bm{p}_{<\mu}\right\}
\right)}
{\sum_{p^\prime_\mu=0}^{V-1}\mathrm{Pr}\left(
\{\bm{P}_{\ge\mu}=\bm{p_{\ge\mu}^\prime}\}\cap
\left\{\bm{R}=\bm{r}\right\}|
\left\{\bm{P}_{<\mu}=\bm{p}_{<\mu}\right\}
\right)}.
\end{eqnarray}

Accordingly, the ``prompt-only \texttt{contextual entropy}'' $S^{(P)}_\mu$ and ``prompt+response \texttt{contextual entropy}'' $S^{(P+R)}_\mu$ are respectively given by
\begin{eqnarray}
S^{(P)}_\mu&\equiv& S\left(\bm{q}^{(\bm{P})}_\mu\right),\\
S^{(P+R)}_\mu&\equiv& S\left(\bm{q}^{(\bm{P+R})}_\mu\right).
\end{eqnarray}

In numerical experiments, we occasionally see examples of large changes between $S^{(P)}_\mu$ and $S^{(P+R)}_\mu$.  It is therefore useful to also construct a divergence measure for prompt token probabilities given the two contexts. We therefore define the Kullback-Leibler divergences
\begin{equation}
\mathrm{KL_\mu}\equiv\mathrm{KL}\left(\bm{q}^{(P)}_\mu||\bm{q}^{(P+R)}_\mu\right)=\sum_{p_\mu=0}^{V-1}
\left[\bm{q}^{(P)}_\mu\right]_{p_\mu}\log\frac{\left[\bm{q}^{(P)}_\mu\right]_{p_\mu}}{\left[\bm{q}^{(P+R)}_\mu\right]_{p_\mu}}.
\label{eq:KL}
\end{equation}
\section{Experiments}
\label{sec:experi}

In this section, we compute attribution scores (\texttt{\texttt{AS}}) for $7$ prompts (\autoref{tab:prompt-desc}) and $8$ LLMs (\autoref{tab:llm-desc}), and evaluate the results. We currently focus on NLP text generation tasks and use greedy and top-p \citep[nucleus,][]{holtzman2020curiouscaseneuraltext} decoding.  
These choices are for the sake of simplicity only, as our method is not limited to the text generation tasks or the decoding strategy. 
Note that these sampling strategies only apply to the initial generation of a response corresponding to a model.  When we compute \texttt{AS} or \texttt{contextual entropy}, we do not sample from the LLM's conditional probability distributions, but rather coerce the model to include the same tokens from the original response into its context as it generates the log-probabilities for each new token.

In each experiment we examine a single response to each prompt from each model. For top-p sampling, responses are selected by identifying the most frequent response generated for a given prompt when the model is prompted 500 times.
We chose $7$ prompts from popular LLM benchmarks~\cite{wang2019gluemultitaskbenchmarkanalysis,
10.1162/tacl_a_00276,
paperno2016lambadadatasetwordprediction,
williams2018broadcoveragechallengecorpussentence, wang2020supergluestickierbenchmarkgeneralpurpose} and modified them while still fulfilling the criteria of the benchmark. For example, we varied the location of the context (beginning, spread out, or end of the sentence), and introduced confusing prompts, or deceptive prompts.

The full results of these experiments are reported in the supplement. In this section, we highlight the results that we find most interesting and relevant.

\noindent\setlength\tabcolsep{4pt}%
\begin{table}[h] 
\captionof{table}{List of Prompts}
\label{tab:prompt-desc}
\centering
\tiny
\begin{tabularx}{\linewidth}{|l|X|}
  \hline
  Prompt ID & Input Prompt \\ [0.1ex]
  \hline
1 & Context:``Tim was nervous about his talk. He had been preparing and practicing for weeks, but he still wasn't sure he was fully ready." Sentence: ``As Tim stood in front of the audience, he took deep breaths and tried to remain\\
\hline
2 & The surfer woke up early to go surfing. He checked the weather forecast to see if it would be windy, since the previous days the weather had been calm. He quickly grabbed his equipment and headed to the beach, hoping the weather would \\
\hline
3 & Dr. Chang was excited about her hike to the mountains. She packed her tent, hiking boots, a warm jacket, and food for the journey. When she reached the mountain top, the view was breathtaking, and Chang felt a sense of \\
\hline
4 & The police officer walked into the room, carefully examining every clue. The information was sparse, but he believed the clues would lead him to the truth. He looked at the doorknob, noticing something odd about it. He made a note to check it more \\
\hline
5 & Passage: ``In 1997, NASA's Cassini space probe was launched to explore Saturn. It was active for twenty years, spending it's last 13 years orbiting Saturn." Question: ``When was the Cassini launched?" \\
\hline
6 & Sentence: ``The water bottle didn't fit into the bag because it was too big." Question: ``What is `it' referring to?" \\
\hline
7 & What year did Albert Einstein visit Mars? \\
  
  \hline
\end{tabularx}
\end{table}

\noindent\setlength\tabcolsep{4pt}%
\begin{table}[h] 
\captionof{table}{Overview of Large Language Models}
\label{tab:llm-desc}
\centering
\tiny
\begin{tabularx}{\linewidth}{|l|c|*{3}{>{\RaggedRight\arraybackslash}X|}}
  \hline
  \textbf{MODEL} & \textbf{Parameters} & \textbf{Description} \\ [0.1ex]
  \hline
gemma-3-1b-it & $1B$ &  Multimodal, text and image input, text output \\ 
\hline
gpt2 & $137M$ & Transformer model, text input, text output \\ 
\hline
gpt-neo-1.3B  & $1.37B$ &  Transformer model, text input, text output, replication of the GPT-3 \\ 
\hline
Llama-3.2-1B-Instruct & $1.24B$ & Multilingual LLM instruction-tuned generative model with text in and text out \\ 
\hline
OLMo-2-0425-1B  & $1.1B$ & Open language model, text input, text output \\ 
\hline
OLMo-2-0425-1B-Instruct  & $1.1B$ & Open language model, text input, text output \\  \hline
Qwen2.5-1.5B-Instruct & $1.54B$ & Instruction-based LLM for generating long texts, coding and mathematics\\
\hline
Qwen3-1.7B & $1.7B$ & Instruction-based LLM for generating long texts, coding, and mathematics. Mixture-of-experts, with agentic capabilities\\
  \hline
\end{tabularx}
\end{table}

\begin{figure}[h]
    \centering
    \includegraphics[width=\linewidth]{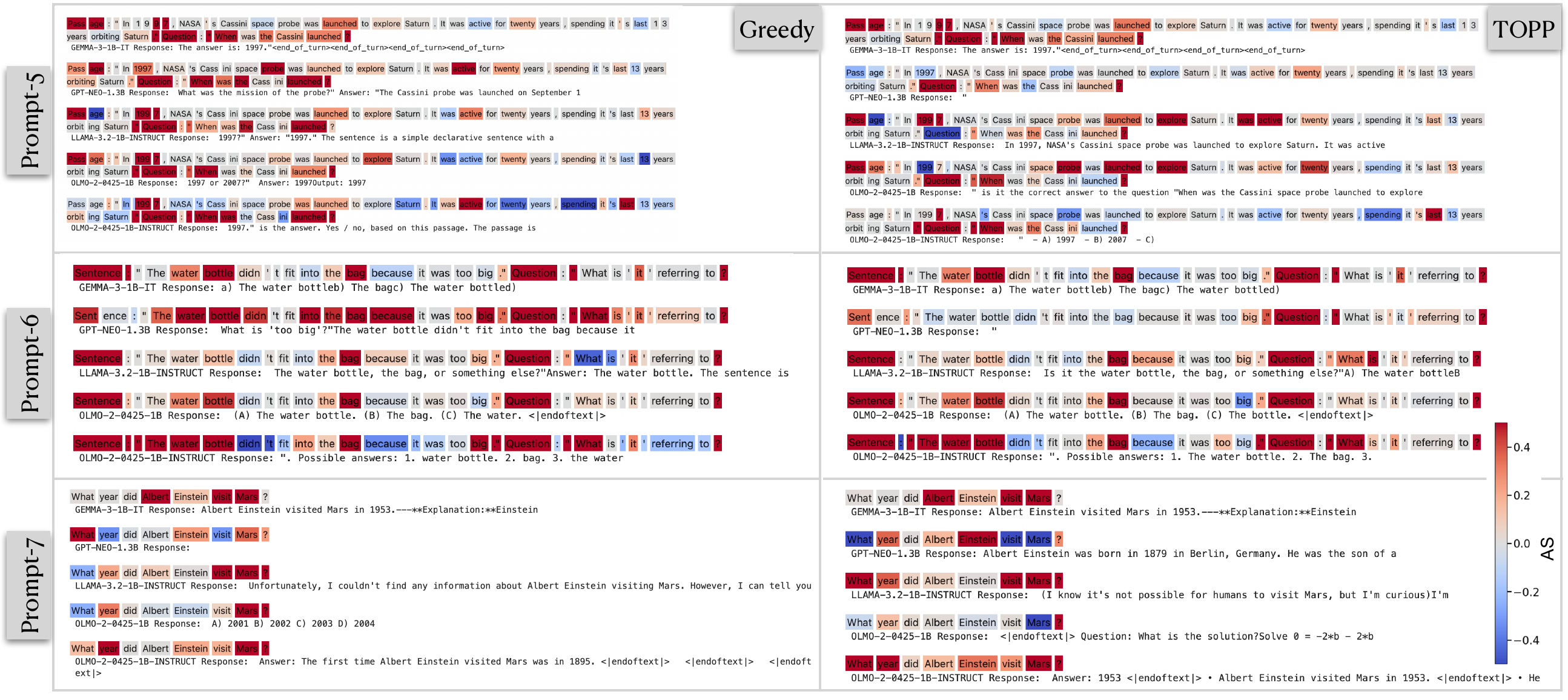}
    \caption{Attribution Scores for prompts given the response for the $5$ LLMs with greedy (left column) and top-p decoding (right column). 
    }
    \label{fig:allenai-asc}
\end{figure}

\subsection{ Attribution Score (AS)}
~\autoref{fig:allenai-asc} shows the  Attribution Scores for 
5 LLMs and 3 prompts. The first column of ~\autoref{fig:allenai-asc} shows greedy and the second shows top-p decoding schemes. As we discuss further below, negative values of \texttt{AS} tell us that there are tokens in the LLM vocabulary that the LLM prefers to the token that the user chose during prompting given the LLM response. In the second row, where the prompt is structured as “Sentence” and “Question”, we see that the LLMs assign high scores to these tokens. We also note that each model gives a reasonable response. An exception to this is \texttt{gpt-neo}, where the default tokenization scheme splits the word ``Sentence" into two tokens, and the attribution is not as high as in the case of other LLMs that do not break the word into subwords. In the first row, the prompt is structured as “Passage” and “Question,” with the answer being a number. As expected, the responses are varied. The models do consider the prompt's structure when devising a response. Here, each model splits the word signifying the prompt structure, “Passage”, into two subwords, and the attribution score is lower, while the word “Question”, tokenized as-is, is assigned a high value of \texttt{AS}. In row 3, an obviously 
misleading prompt is used to invoke an LLM response. With the exception of \texttt{Llama}, we see that the LLMs using the greedy and top-p decoding strategies provide incorrect responses. Note that \texttt{gpt-neo} responds with multiple blank spaces for greedy decoding.
 
\subsection{Contextual Entropy and KL Divergence  of Token Distributions Conditioned on the Remaining Context}
\begin{figure}[h]
    \centering
    \includegraphics[width=\textwidth]{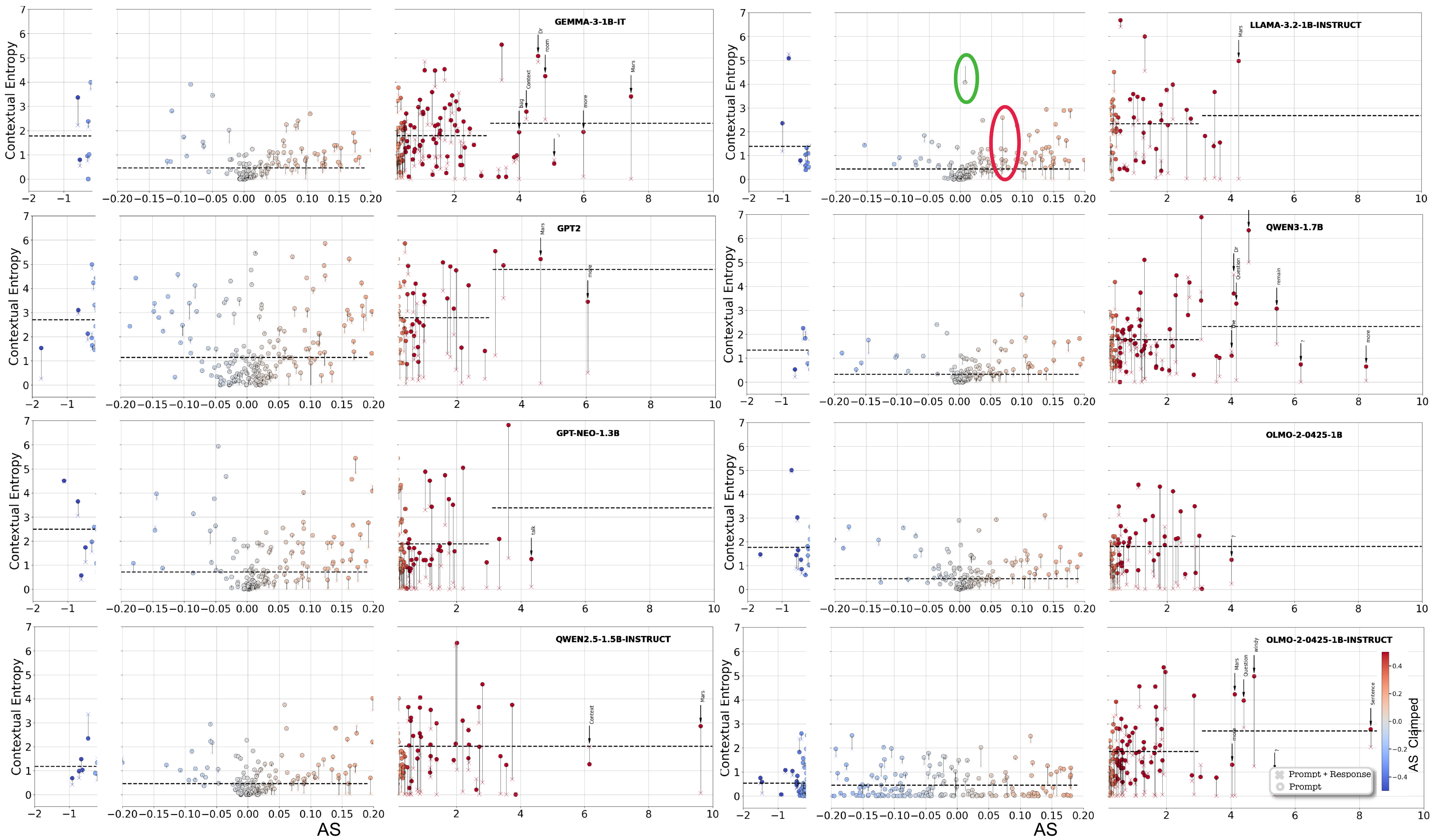}
    \caption{Contextual Entropy versus Attribution Scores for seven prompts and $8$ LLMs and a~\textit{greedy} response. The x-axis is stretched from $-0.2$ to $0.2$ for clarity. The entropy is computed with (\texttt{$\times$}) and without (\texttt{o}) including the response. The dashed lines indicate the mean of the entropy ($S^{(P)}$) in the bucket. The green and red circles are the anomalies discussed in Subsection \ref{sec:anomaly}}
    \label{fig:allenai-asc-2-greedy}
\end{figure}

\begin{figure}[h]
    \centering
    \includegraphics[width=\textwidth]{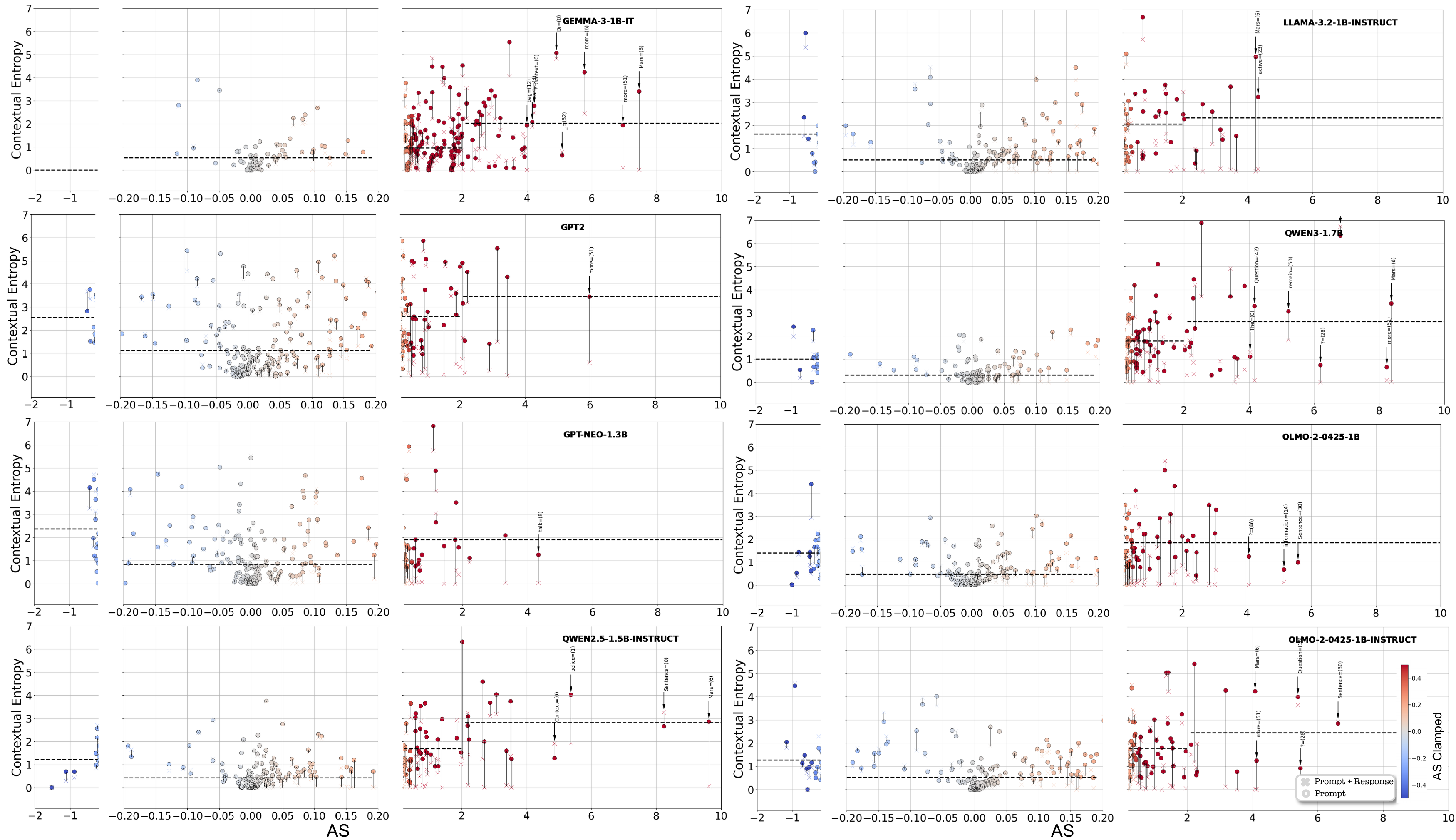}
    \caption{Contextual Entropy versus Attribution Scores for six prompts and $8$ LLMs and a response downsampling strategy using~\textit{top-p}. The x-axis is stretched from $-0.2$ to $0.2$ for clarity. The entropy is computed with (\texttt{$\times$}) and without (\texttt{o}) including the response. The dashed lines indicate the mean of the entropy ($S^{(P)}$) in the bucket.}
    \label{fig:allenai-asc-2-topp}
\end{figure}

~\autoref{fig:allenai-asc-2-greedy} and~\autoref{fig:allenai-asc-2-topp} show~\texttt{contextual entropy} versus \texttt{AS}, where \texttt{AS} is ordered by value, for $7$ prompts and $8$ LLMs with a response downsampling strategy of greedy and top-p respectively. 
The \texttt{y-axis} shows the contextual entropies $S^{(P)}$ and $S^{(P+R)}$ of the distribution of each token.
The figures represent $S^{(P)}$ by $\circ$ and $S^{(P+R)}$ by $\times$.
In each panel, the \texttt{AS} axis is divided up into three ranges: high, near-zero, and negative, with breaks in the axis between ranges so as to permit convenient representation of the full dynamic range.
The mean $S^{(P)}$ in each range is indicated by the horizontal dashed lines.
A low contextual entropy means that given the context ($\circ$ or $\times$) the model prefers fewer replacement tokens for the given token. These low contextual entropies are mostly concentrated around near-zero \texttt{AS}, also seen in near-zero \autoref{fig:allenai-asc} (gray).
Similarly, high contextual entropy corresponds to more uncertainty in potential token replacements candidates. These high contextual entropies are mostly concentrated in positive and negative \texttt{AS}, also seen in blue and red hues in \autoref{fig:allenai-asc}.

From ~\autoref{fig:allenai-asc-2-greedy} and ~\autoref{fig:allenai-asc-2-topp}, near-zero \texttt{AS} corresponds to low contextual entropies ($\circ$ \& $\times$), with a few outliers. When the $\circ$ \& $\times$ overlap, the context of the response does not change the entropy for a token distribution conditioned on the given context. In such cases we found that the original prompt token was one of the top $90\%$ candidates for token replacement, further discussed in~\autoref{fig:entropy-tokens}.
In cases when the \texttt{AS} is negative, the numerator of~\autoref{eq:Attribution} is less than the denominator. This means the model prefers a token other than the current prompt token for the model original response. However, the original token may still be a part of the top replacement token candidates, further discussed in~\autoref{fig:entropy-tokens}-b.

\subsubsection{Training Convergence}
The ~\texttt{contextual entropy} 
($S^{(P)}$ and $S^{(P+R)}$)
shows how adding the response to the conditioning information affects the uncertainty in the potential replacements for the token. Our attribution score together with the ~\texttt{contextual entropy} allows us to infer how well the model converges onto the distribution of the training data. The evidence for this is seen in ~\autoref{fig:allenai-asc-2-greedy} and ~\autoref{fig:allenai-asc-2-topp}. In the figures, we see that almost always this uncertainty is reduced with the addition of response, and this dip in entropy is larger for higher \texttt{AS} values. The entropy is low for \texttt{AS} values that are in the near-zero range ($-0.1$ to $0.1$), except for older models like \texttt{gpt2} and \texttt{gpt-neo}.~\texttt{gpt2} is an older English-only causal LLM. \texttt{gpt-neo} is newer than \texttt{gpt2}, trained on the Pile,~\citep{gao2020pile,radford2019language}. Its model card reports better results than \texttt{gpt2} on several language benchmarks, such as LAMBADA~\citep{zellers2019hellaswagmachinereallyfinish} and HellaSwag~\citep{paperno2016lambadadatasetwordprediction}. However, both are clearly behind compared to newer models. In contrast, modern instruct-tuned models such as Llama 3.1 Instruct are designed for assistant-style dialogue, multilingual use, code, and long-context work.

The larger \texttt{contextual entropy} at near-zero \texttt{AS} values exhibited by the \texttt{gpt2} and \texttt{gpt-neo} models indicate high uncertainty in replacement token choice. 
For a model whose training has converged and therefore has a well ascertained distribution over texts, this would be anomalous behavior, because given a ``reasonable'' prompt (i.e. one consistent with the training data distribution) a prompt token with near-zero \texttt{AS} should correspond to a situation where the identity of that token is implied by the remaining context i.e. it can be easily guessed given the rest of the sentence, so that its \texttt{contextual entropy} should be low.  If instead such a token has high \texttt{contextual entropy}, this means that it is a pure noise token, contributing nothing to the probability of the response (near-zero \texttt{AS}) while being largely unconstrained by the context (high \texttt{contextual entropy}). Such noise tokens should be rare for reasonable prompts to a model with good training convergence. We therefore interpret a high frequency of such noise tokens as indication of poor model training convergence.

Negative values of \texttt{AS} tell us that there are tokens in the LLM vocabulary that the LLM prefers to the token that the user chose during prompting given the LLM response [\textbf{C3}]. The ~\texttt{contextual entropy} for negative \texttt{AS} gets higher as we move towards more negative range, indicating that the model finds many other tokens more likely. 

\subsubsection{Investigating Anomalies and Prompt Analysis}
\label{sec:anomaly}

\begin{figure}[h]
    \centering
    \includegraphics[height=0.35\linewidth,width=\textwidth]{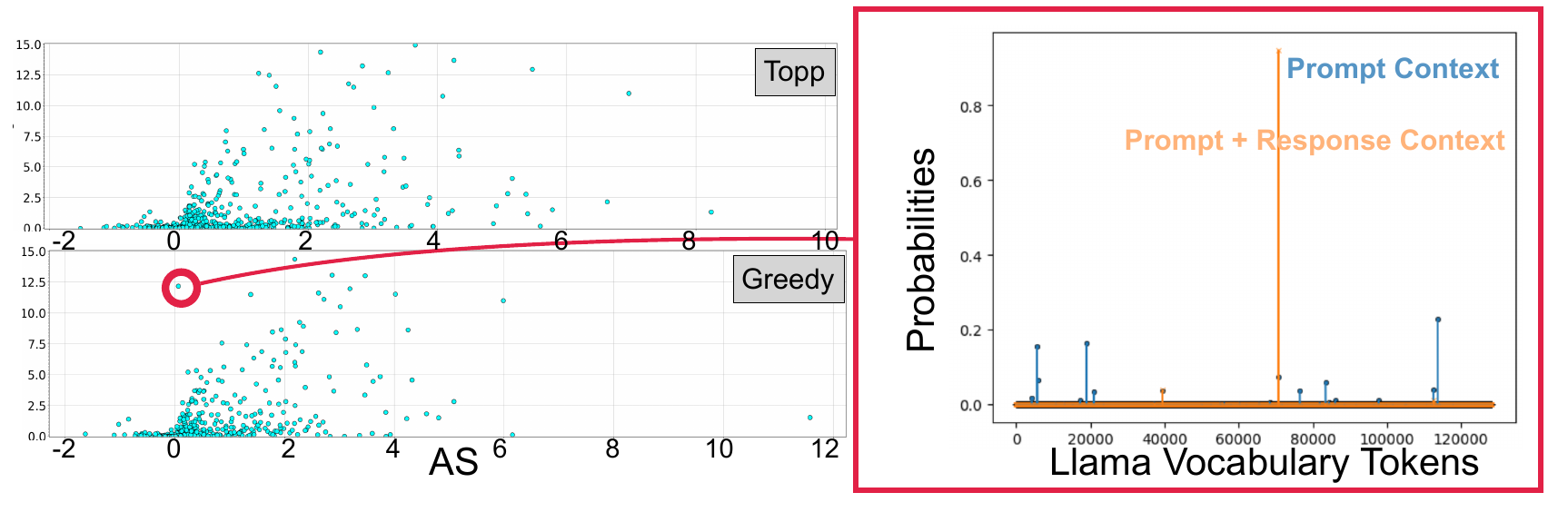}
    \caption{KL divergences defined in Equation (15) for greedy and top-p sampling.}
    \label{fig:KL}
\end{figure}

\begin{figure}[h]
    \centering
    \includegraphics[width=\textwidth]{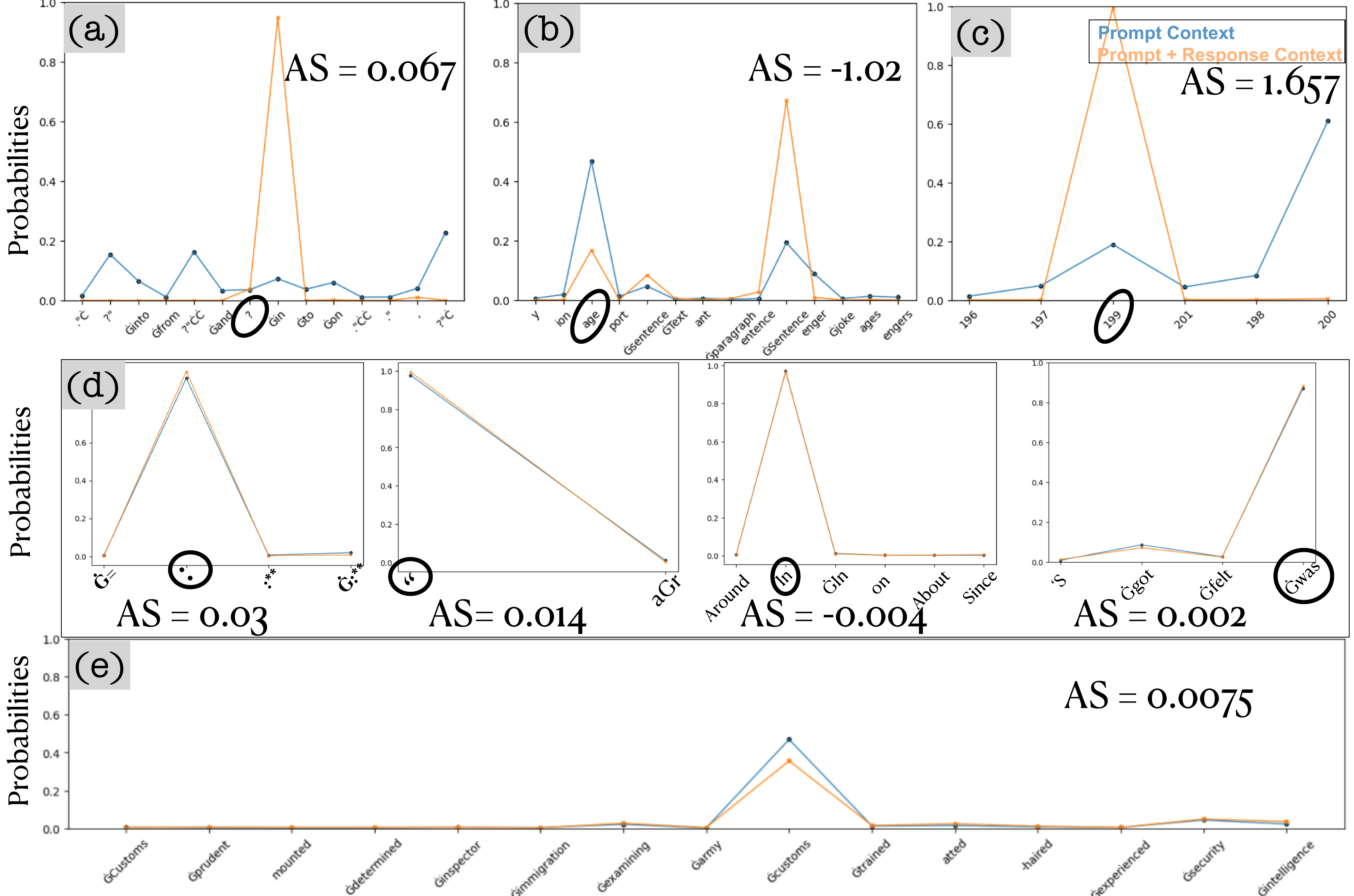}
    \caption{Token probabilities used in the computation of~\texttt{contextual entropy} with and without the response context. The prompt tokens are highlighted in the circles. (a) shows the token highlighted with a red oval in~\autoref{fig:allenai-asc-2-greedy} and the candidate replacements the~\texttt{llama-3.2-1B-Instruct} prefers. (b) shows a token with negative~\texttt{AS}. (c) shows tokens with large-positive~\texttt{AS}, (d) (e) shows tokens with near\-zero~\texttt{AS}. Note: $\dot{G}$ represents and empty space. }
    \label{fig:entropy-tokens}
\end{figure}

Figure~\ref{fig:KL} shows the KL divergences
between the token distributions with and without the response context, defined in Equation (\ref{eq:KL}), for greedy and top-p sampling. ~\autoref{fig:entropy-tokens} shows the token probabilities used in the computation of~\texttt{contextual entropy} with and without the response context, with prompt tokens highlighted in the black circles.~\autoref{fig:entropy-tokens}-(a) shows the token highlighted with a red oval in~\autoref{fig:allenai-asc-2-greedy} and the candidate replacements the~\texttt{llama-3.2-1B-Instruct} prefers.~\autoref{fig:entropy-tokens}-(b) shows a token with negative~\texttt{AS}.~\autoref{fig:entropy-tokens}-(d) shows the token probabilities with near-zero~\texttt{AS}.

~\autoref{fig:entropy-tokens}-(d), where the token probabilities with near-zero~\texttt{AS} are shown, reveals that the candidate replacements for prompt tokens are unchanged even when the response is introduced into the context. For a negative and positive ~\texttt{AS},~\autoref{fig:entropy-tokens}-(b) and -(c) respectively, the log-probabilities change for multiple tokens, including the original token, which is reflected in the ~\texttt{AS} scoring as these are the more sensitive tokens in the prompt.

We now demonstrate how these kinds of plots can be used to detect, investigate, and intervene upon anomalies in LLM output.

\paragraph{Anomaly 1:} As shown in Figure~\ref{fig:KL}, the KL divergence is mostly insignificant for near-zero \texttt{AS} values and is higher otherwise. 
One exception in the near-zero \texttt{AS} range with KL divergence value of $12.15$ is in greedy sampling and is highlighted with a red oval in~\autoref{fig:allenai-asc-2-greedy} and a red circle in~\autoref{fig:KL}. Upon further investigation we found that this token, (the final ``\textbf{$?$}'' in \textbf{When was the Cassini launched\textit{?}}), belongs to prompt 5 (\autoref{tab:prompt-desc}) for the~\texttt{llama-3.2-1B-Instruct} model. This is also a token where the prompt$+$response ~\texttt{contextual entropy} is much lower than prompt-only ~\texttt{contextual entropy} (see~\autoref{fig:allenai-asc-2-greedy}) which is unusual at near-zero \texttt{AS}. ~\autoref{fig:KL} (outlined in red) shows that the model with prompt+response context has fewer candidate replacements ~\textbf{`:', `?', ` in', ` on', `,'} which include the original prompt token, \textbf(?). This exception is also shown in~\autoref{fig:entropy-tokens}-(a), where the model prefers the token ``\textit{ in}" over the original prompt token, ``\textit{?}". The reason lies in the response, ``~\textit{1997?" Answer: ``1997." The sentence is a simple declarative sentence with a}". The models choice of the prompt token, ``\textit{ in}", with response context included makes the sentence more cohesive (``\textit{When was the Cassini launched in 1997?..."}) than with the original prompt, ``\textit{When was the Cassini launched? 1997?...}". 
In effect, the model would have been ``happier'' with a slight rephrasing of the question!

\paragraph{Anomaly 2:}
Another exception is highlighted in green oval in~\autoref{fig:allenai-asc-2-greedy}, in the near-zero \texttt{AS} range. The anomaly consists of the fact that, unusually, $S^{(P+R)}>S^{(P)}$ for this token. In effect, the information added by the response over that in the prompt alone has broadened the distribution for this token. Note that, typically, the token probabilities with near-zero \texttt{AS} (refer to \autoref{fig:entropy-tokens}-(d)) reveal that the candidate replacements for prompt tokens remain unchanged even when the response is introduced into the context. 

The anomaly corresponds to prompt 4 with a response:
``\textit{closely. 
He walked over to the door and examined the doorknob more closely. It was a…}". The prompt token in question is ``~\textbf{police}" (at position 1). Upon further investigation the we found that the~\texttt{llama-3.2-1B-Instruct} model has a substantially higher preference for the tokens ``~\textbf{customs}" and ``~\textbf{security}" than for the original prompt token ``~\textbf{police}", refer ~\autoref{fig:entropy-tokens}-(d). Here, we show tokens whose token probabilities are greater then $0.005$.
This usage aligns more closely with terminology commonly found in US English, as opposed to UK English or other English variants. Using these examples, we can gain insights into the type of dataset used to  train the LLMs and how well the model converges to this training data.

\paragraph{Anomaly 3:}
~\autoref{fig:entropy-tokens}-(b) shows token ``\texttt{age}", from ``Pass\textit{age}" at position 1, with negative~\texttt{AS}, belonging to prompt-5 and ~\texttt{llama-3.2-1B-Instruct} model, refer~\autoref{fig:allenai-asc} for greedy sampling. The other candidate token the model prefers is ``\textit{entence}", which could be seen as a subword of the word ``\textit{Sentence}. Upon prompting the model with the word ``\textit{Passage}", at position 1, replaced with ``\textit{Sentence}", we were able to get a new, more coherent response: ``\textit{The answer is 1997. The sentence is in the past tense, and the subject is...}". The original response was ``\textit{1997?" Answer: ``1997." The sentence is a simple declarative sentence with a}".
Interestingly, when we only replaced the token ``\textit{age}" with ``\textit{entence}" (giving us ``\textit{Passentence}") we obtained the same coherent response. This example shows that we are able to identify causes of poor output, and engineer prompts to produce improved responses.
At the same time we gain an improved understanding of the nature of the model's training dataset [\textbf{C3}], in that the ``Passage/Question'' format appears to have been unusual in the dataset, resulting in confused output. 

\begin{figure}[h]
    \centering
    \includegraphics[width=\textwidth]{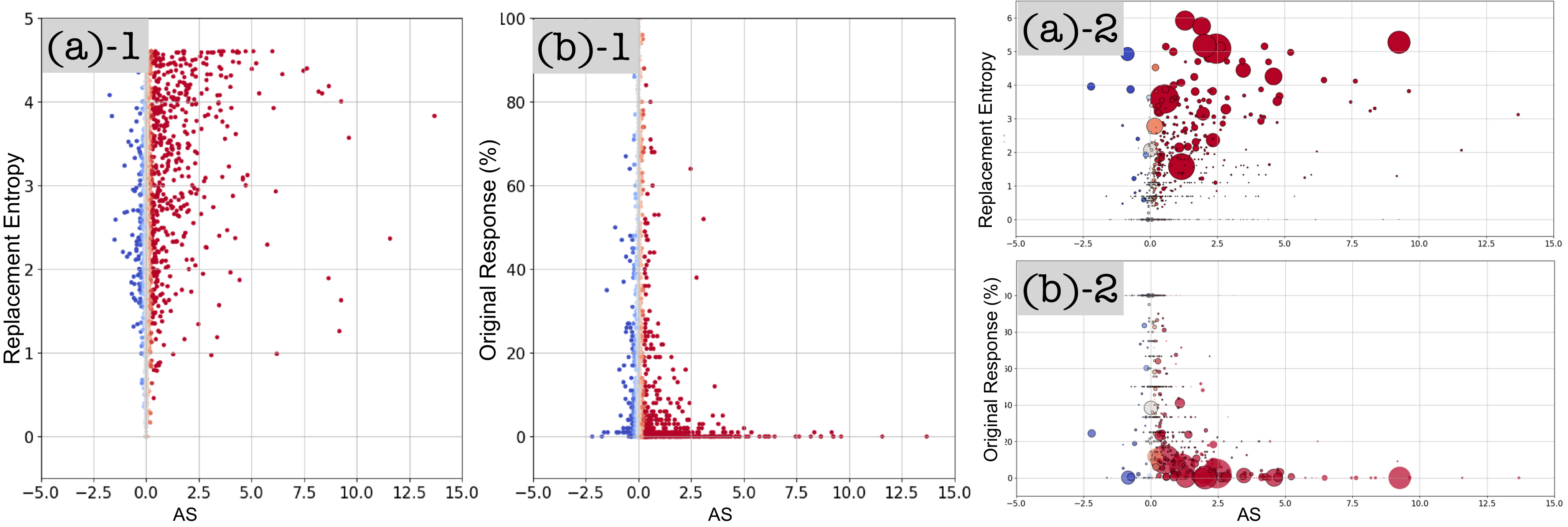}
    \caption{Figure (a) shows the entropy of the frequency counts of exact matches in the responses for both greedy and top-p sampling versus \texttt{AS}. The frequency counts are computed for each token replacement with the top 90\% tokens. Each circle is one token. (b) shows the percentage of original response for each token versus \texttt{AS}. In (a,b)-2, the size of the circles indicate the number of top $90\%$ tokens used to replace each token.}
    \label{fig:logasent2}
\end{figure}

\subsection{Replacement Entropy  of Frequency Counts of the Exact Response Matches} 
So far, we have computed \texttt{AS} for a single response; here, we will examine how  \texttt{AS} is related to the preferred responses of an LLM.
For Figure~\ref{fig:logasent2}-(a) we compute the frequency counts of responses for each token replacement with the top $90\%$ tokens. To compute the “top $90\%$ tokens,” the LLM output are converted to probabilities, sorted in descending order, and the top tokens are selected whose cumulative probability mass reaches $90\%$.  First we take the top $90\%$ candidate tokens for each token in the prompt. Then we prompt the model with the candidate tokens in place of the original token, in each case leaving all the other tokens unchanged. This results in multiple responses. We do this for greedy and top-p sampling strategies.
We then compute the frequency counts of responses (where all response tokens must match for responses to be counted as the same). Normalized, this frequency distribution becomes an empirical probability distribution.  We compute the entropy of this distribution, and refer to it as the  ~\texttt{replacement entropy}. We also tally the number of exact matches to the original response, for both greedy and top-p sampling strategies.

Figure~\ref{fig:logasent2}-(a), shows \texttt{replacement entropy} versus \texttt{AS}. The size of the circles on the plots on the right indicates the number of top 90\% tokens used to replace each token (each circle is one token) for both greedy and top-p strategies.  As the \texttt{AS} increases (or goes more negative) the variability in responses increase even when there are fewer candidate replacement tokens. When \texttt{AS} is near zero, there are fewer top 90\% replacement tokens and relatively lower~\texttt{replacement entropy}. 
Figure~\ref{fig:logasent2}-(b) shows the percentage of original response for each token versus \texttt{AS}. The large the value of \texttt{AS} the lower is the original response count even with larger number of candidate replacements when compared to near-zero \texttt{AS}. We can conclude that the high~\texttt{AS} tokens are more sensitive, have more variation in the responses, and do not prefer the original response even when the top $90\%$ candidate tokens are chosen.  Unlike ~\texttt{contextual entropy}, which computes the entropy from log-probabilities extracted from the models,~\texttt{replacement entropy}  uses the frequency counts of the responses and we can see a clear correlation between \texttt{AS} and the LLM response across multiple models and prompts [\textbf{C1, C3}].

\section{Evaluation}
\label{sec:eval}

\begin{figure}[h]
    \centering
    \includegraphics[width=\textwidth]{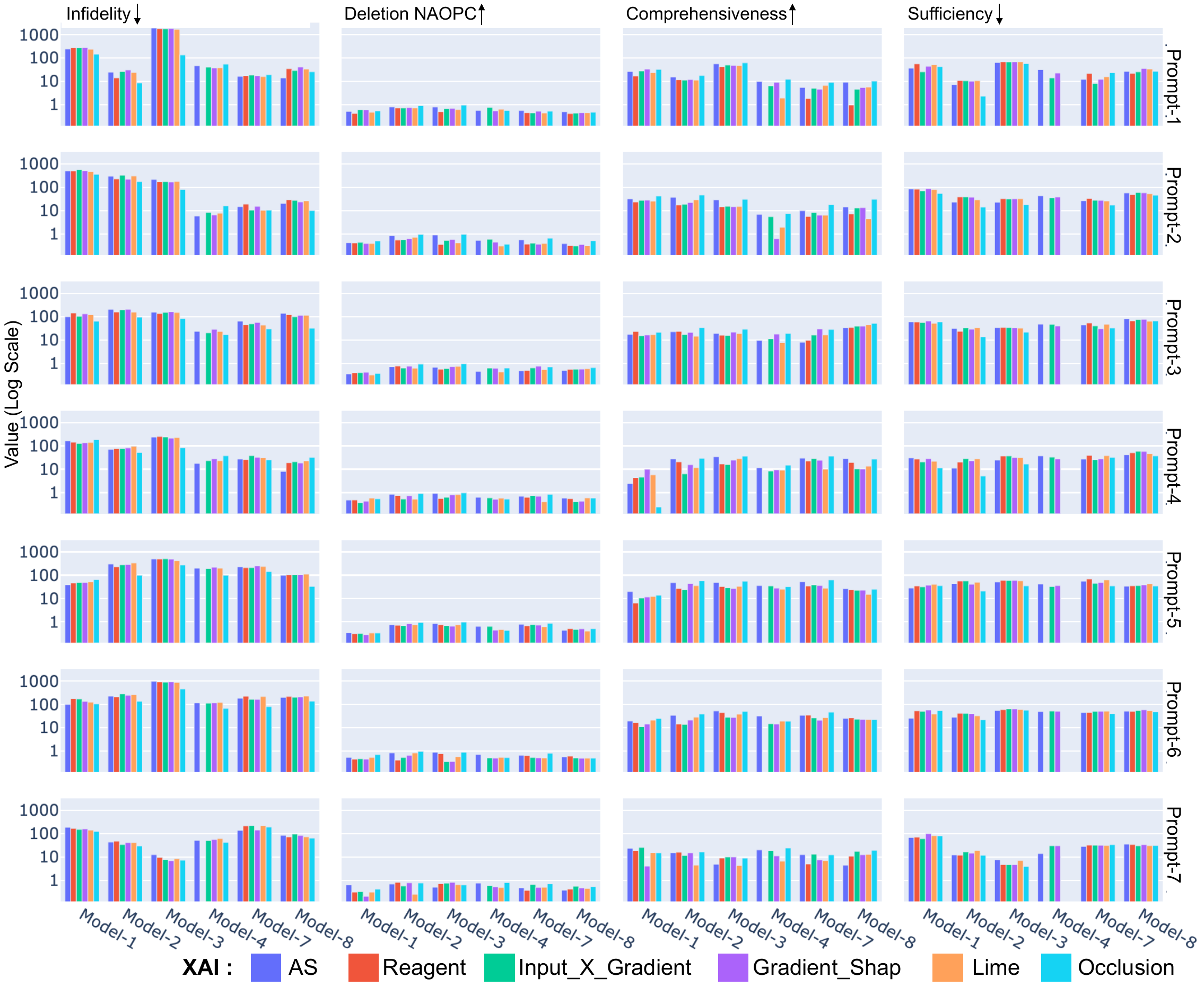}
    \caption{XAI Evaluation Metrics for $6$ models and $7$ prompts for greedy sampling. Model $1$ = gemma-3-1b-it, $2$ = gpt2, $3$ = gpt-neo-1.3B, $4$ = Llama-3.2-1B-Instruct, 
    $7$ = Qwen2.5-1.5B-Instruct, and $8$ = Qwen3-1.7B.}
    \label{fig:logas_eval_greedy}
\end{figure}

\begin{figure}[h]
    \centering
    \includegraphics[width=\textwidth]{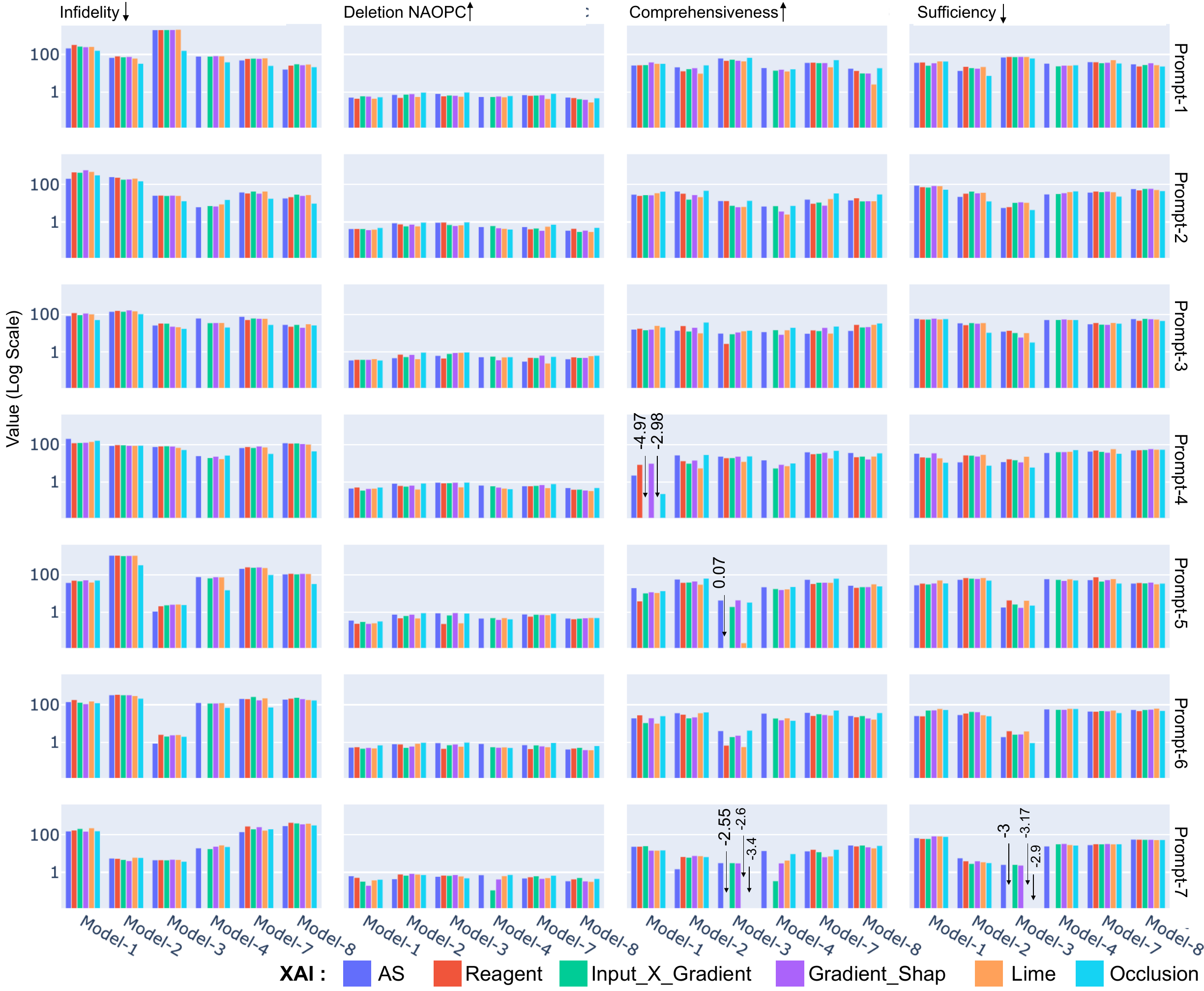}
    \caption{XAI Evaluation Metrics for $6$ models and $7$ prompts for top-p sampling. Model $1$ = gemma-3-1b-it, $2$ = gpt2, $3$ = gpt-neo-1.3B, $4$ = Llama-3.2-1B-Instruct, 
    $7$ = Qwen2.5-1.5B-Instruct, and $8$ = Qwen3-1.7B. }
    \label{fig:logas_eval_topp}
\end{figure}
We evaluate our token-attribution methods across $6$ LLMs, given a prompt, $x$, and a fixed response, $y$. We test whether the attribution scores are faithful to the model’s actual behavior and compare it with other XAI methods: \texttt{Input\_X\_Gradient}~\citep{DBLP:journals/corr/ShrikumarGSK16}, \texttt{Gradient\_Shapley}~\citep{lundberg2017shap}, \texttt{Lime}~\citep{DBLP:journals/corr/RibeiroSG16}, \texttt{Occlusion}~\citep{DBLP:journals/corr/ZeilerF13}, and \texttt{ReAgent},~\cite{zhao2024reagentmodelagnosticfeatureattribution}. 

\texttt{Input\_X\_Gradient} ($input\_x\_gradient$) is a local, first-order attribution technique that assigns token-level importance by taking the element-wise product of the input representation (typically token embeddings) and the gradient of a chosen target quantity (e.g., a  log-probability) with respect to that input. This yields a saliency score that approximates the token’s marginal influence under an infinitesimal perturbation assumption. \texttt{Gradient\_Shapley} (often implemented via Shapley-inspired gradient methods such as $gradient\_shap$) instead estimates token contributions by aggregating gradient-based attributions across multiple interpolations between a baseline and the observed input, thereby approximating Shapley-style credit assignment and typically reducing sensitivity to the local gradient at the input point.
\texttt{Occlusion} is a perturbation-based attribution method that quantifies token importance by systematically masking (or replacing) each token or token span and measuring the induced change in a specified target score, such as the log-probability of a generated token. By comparing these score differences to the baseline, it estimates each input region’s contribution. Occlusion works without access to internal gradients.
\texttt{LIME} (Local Interpretable Model-agnostic Explanations) is another perturbation-based method that samples many locally perturbed variants of an input (e.g., by deleting or masking tokens), but it fits a simple, interpretable surrogate model to approximate the black-box model’s behavior in that neighborhood. The surrogate’s coefficients are then interpreted as token-level attributions, providing a locally faithful explanation of which input components most influence the target prediction.
\texttt{ReAgent} is a perturbation-based, model-agnostic method for autoregressive generation that scores each context token by replacing it with plausible alternatives and measuring the resulting change in a target quantity (e.g., the next-token log-probability). By repeating these replacement probes (replacing\_ratio and num\_probes) and aggregating the induced score drops, it identifies the minimal subset of tokens that most strongly sustains the model’s prediction.

We apply $4$ evaluation methods to test the faithfulness of the XAI methods to the model’s actual behavior. The evaluation methods used are Infidelity \citep{yeh2019infidelitysensitivityexplanations}, Deletion/NAOPC curves \citep{edin2025normalizedaopcfixingmisleading}, Comprehensiveness \citep{deyoung2020eraser}, and Sufficiency \citep{deyoung2020eraser}. Infidelity measures how well an explanation’s attributions predict the actual change in a chosen model $f$ when the input is randomly perturbed. For each perturbation, $x'$, we compare the score change $f(x)-f(x')$ to the change the explanation predicts (often a dot product between attributions and the perturbation mask), then average the squared error across many perturbations. Lower infidelity means the explanation is more faithful to the model under that perturbation distribution, where $0$ would be a perfect prediction. The absolute magnitude depends on the scale of $f$ and the perturbation strength, so it’s most interpretable when comparing methods under the same setup. Deletion evaluates faithfulness by ranking tokens by attribution and progressively ``removing" or neutralizing the top-k tokens, then recomputing the model score $f$ (often $log(P(y|x))$) for a fixed response). A good explanation causes the score to drop quickly when its highest-ranked tokens are deleted. 
NAOPC (Area Over the Perturbation Curve) summarizes the score as the average drop $f(x)-f(x(k))$ across deletion steps, and a higher NAOPC indicates that the explainer’s top tokens are more causally important for the model’s behavior. Results depend on the deletion operator (e.g., masking vs replacement) and should be compared under identical perturbations. Here, we use replacement. Comprehensiveness measures whether the tokens the explanation marks as important are truly necessary for the model’s decision. We take the top-k tokens $T_k$ and measure how much the model score falls when we remove them: $f(x)-f(x/T_k)$. Higher comprehensiveness means this removal substantially harms the model score, suggesting the explanation captured key evidence. Because token deletion can introduce a distribution shift, many implementations replace tokens with a baseline to keep the length or tokenization stable. Sufficiency asks whether the selected top-k tokens are nearly enough on their own to support the model’s score. We retain $T_k$ and compute the drop: $f(x)-f(T_k)$. Lower sufficiency (a small drop) indicates the explanation’s subset retains most of the model’s evidence for the output, while a large drop suggests the subset is incomplete. Sufficiency is often reported alongside comprehensiveness to distinguish ``necessary" evidence from ``nearly sufficient" evidence.

We use the Inseq library~\citep{sarti-etal-2023-inseq}, a unified, research-oriented framework for computing and visualizing token-level attributions for modern sequence models, to compute the attribution scores of other XAI methods. We then compare \texttt{AS} with other XAI methods using the evaluation metrics. This approach improves comparability and reproducibility of XAI analyses across models, attribution techniques, and decoding settings, thereby facilitating systematic evaluation and reporting. 
We compare \texttt{AS} against five XAI methods: \texttt{Input\_X\_Gradient}, \texttt{Gradient\_Shapley}, \texttt{Lime}, \texttt{Occlusion}, and \texttt{ReAgent}. 

Figure~\ref{fig:logas_eval_greedy} and Figure~\ref{fig:logas_eval_topp} show the evaluation metrics for $6$ LLMs and $7$ prompts for greedy and top-p sampling strategies, respectively. The direction of the arrow indicates which values are better; for example, up means higher is better, and down means lower is better. In most cases, we see that \texttt{AS} performs either better or within a close range of the other more popular methods. Also, note that while \texttt{AS} is model-agnostic and provides results across multiple LLMs, $input\_x\_gradient$ and $gradient\_shapley$ depend on the model architecture. Note that the \texttt{OLMo} models are not implemented in the current version of Inseq and are not shown in the figures, even though we can compute \texttt{AS} for these models. As of writing the paper, Inseq could not apply \texttt{ReAgent} to the \texttt{Llama} model in our system. All of the XAI implementations require a user-defined set of parameters for computing the attribution scores. For example, ReAgent has $replacement\_ratio$ and $num\_probes$. Hence, the final token attributions depend on the user-defined parameters. By contrast, \texttt{AS} is a parameter-free, model-agnostic attribution method that provides attribution based only on the model's log probabilities for a given prompt and response. With sensitivity built into~\texttt{AS} and with~\texttt{AS} being parameter-free, our method could thus serve as a baseline for hyperparameter selection for other XAI methods [\textbf{C1, C2, C4}].

\begin{figure}[h]
    \centering
    \includegraphics[width=\textwidth]{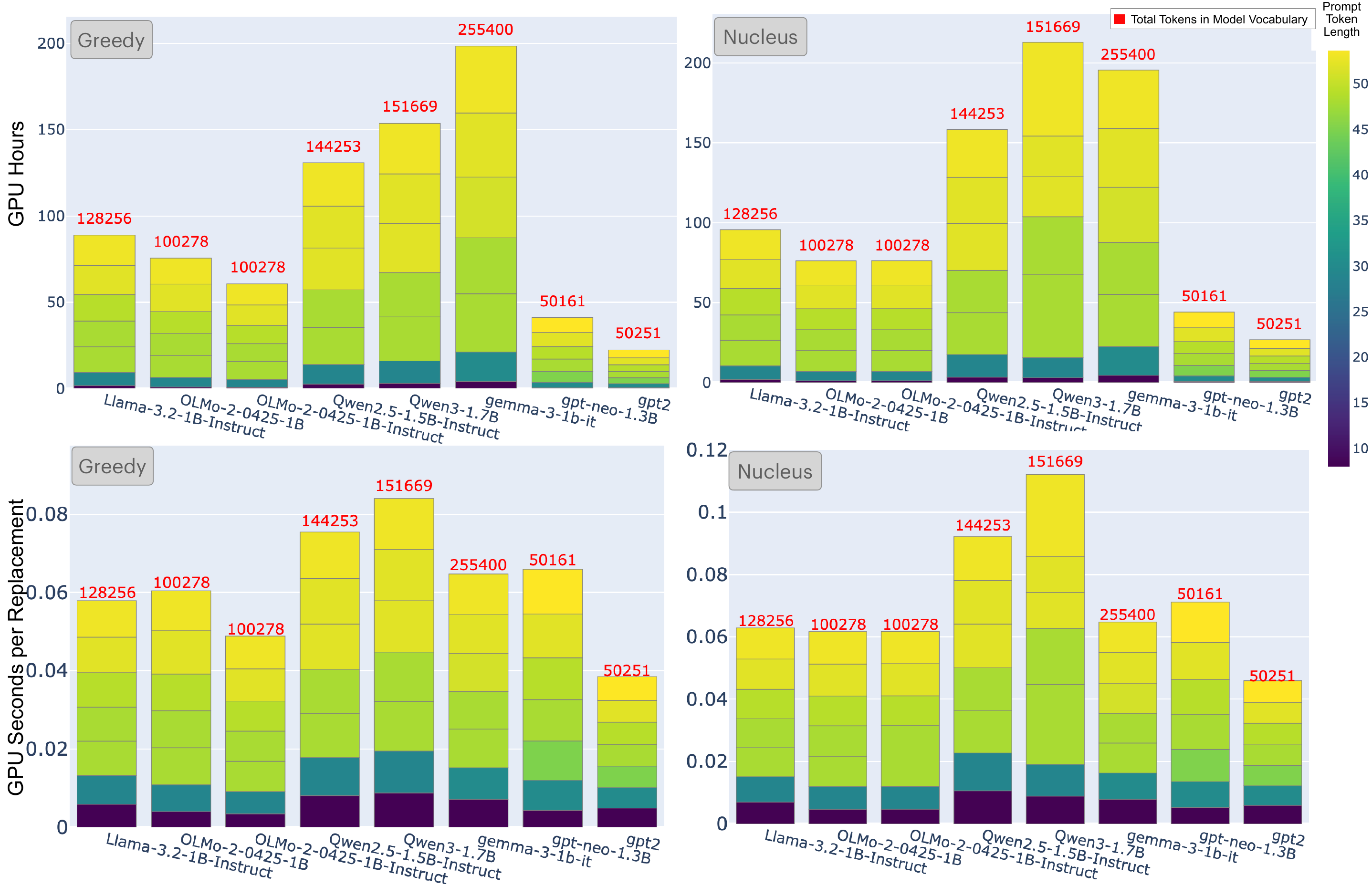}
    \caption{Compute time for~\texttt{AS} for greedy and nucleus (top-p) decoding across multiple LLMs. Top row shows the total GPU hours and bottom row shows GPU seconds for each token replacement from Equation~\ref{eq:Denom}. The red text on each stacked bar is the total tokens in the model vocabulary. The prompt token length is the length of the LLM tokenized prompt.}
    \label{fig:comp_time_greedy_topp}
\end{figure}

~\autoref{fig:comp_time_greedy_topp} shows the compute time for~\texttt{AS} for greedy decoding across multiple LLMs. We show total GPU hours on the top and GPU seconds for each token replacement from Equation~\ref{eq:Denom} at the bottom. The red text on each stacked bar is the total tokens in the model vocabulary. As  the model vocabulary increases, the time to compute the~\texttt{AS} also increases. We use an A100 GPU of a leadership-class supercomputer to compute the results. The goal of this work is to gain a deeper understanding of how large language models produce responses, using a model-agnostic attribution scoring method that makes minimal assumptions. This approach comes at a cost of higher GPU hours. 
Note, however, that it is likely that the \texttt{AS} method can be made more computationally-efficient than it is in our current implementation, by restricting the summations over replacement tokens in Equations (\ref{eq:Attribution_3}--\ref{eq:Attribution_4}) to only the top probability tokens. We will investigate such potential computational savings in future work.

Multiple repeated prompting cause log-probability variation on GPUs, even when the generated text is identical for the same prompt, and this is typically attributable to numerical and algorithmic nondeterminism in accelerated inference. In particular, CUDA/cuDNN kernels and fused attention implementations may employ parallel reductions, atomics, or autotuned kernel selection, yielding small floating-point differences because arithmetic is not strictly associative. These effects are often amplified when precision is reduced (FP16/BF16) or when matrix multiplications are TF32-enabled. Additional contributors include stochastic decoding configurations (e.g., sampling) that may, by coincidence, produce the same token sequence, and inadvertent training-mode behavior (e.g., dropout) if the model is not set to evaluation/inference mode. Consequently, the token choice can remain stable while the underlying log-probabilities shift slightly, unless determinism is enforced via fixed decoding, deterministic algorithm settings, disabled benchmarking/autotuning, and stricter precision controls, which is precisely what we do for~\texttt{AS} computations, since we use GPUs for speedy calculations. The ~\texttt{AS} computations would be expedited if GPU non-determinism did not need to be considered.

In ~\autoref{sec:experi}, we have seen empirical results of promising near-zero AS behavior, which  hints towards the quality of model convergence. A part of our future work is to use this finding to identify model checkpoints to guide the LLM training. Our hypothesis is that this finding could help identify bad training and insufficient model capacity during the training process. 
\section{Conclusion}

We have developed an attribution scoring method, \texttt{AS}, that leverages the generative nature of current Large Language Models (LLMs), expressed by the conditional probabilities they compute to sample each response token given the previous tokens. Our model-agnostic probabilistic token attribution measure 
exploits the connection between LLMs and stochastic processes, situating all LLMs in the context of the well-developed mathematical theory of stochastic processes, and permitting broad cross-comparisons of LLM output across all such models.  We use this framework to reconstruct the probability distribution over texts that LLMs build up during training from the log-probabilities emitted by the models during inference.
From our experiments and evaluations using \texttt{contextual entropy}, \texttt{replacement entropy}, and KL divergence of token distributions conditioned on the entire remaining context, we can draw conclusions about LLM stability. We found that GPT -2 and GPT-Neo are less effective at text generation than newer models like Llama. Despite having no access to the training data, we can draw conclusions about how well the models converge on the training data distribution. We were also able to tweak the prompts with modified tokens to generate a desired response. AS has sensitivity built into its definition and is parameter-free, which could serve as a guide to selecting hyperparameters for other XAI and XAI evaluation methods that are faster compared to our approach. We evaluate 8 models across 7 prompts and identify token sensitivity and response stability, thereby improving interpretability and guiding users to focus on uncertain or unstable parts of the generation.

\acks{
This research used resources of the Argonne Leadership Computing Facility, which is a U.S. Department of Energy Office of Science User Facility operated under contract DE-AC02-06CH11357. All authors were supported by the Office of Science, U.S. Department of Energy, under contract DE-AC02-06CH11357.}

\vskip 0.2in
\bibliography{00_sample}

\appendix
\section{}
\label{app:theorem}

In this appendix, we provide the Attribution score for 7 prompts and 8 models for both greedy and topp sampling.

\begin{figure}[h]
    \centering
    \includegraphics[height=1.4\linewidth]{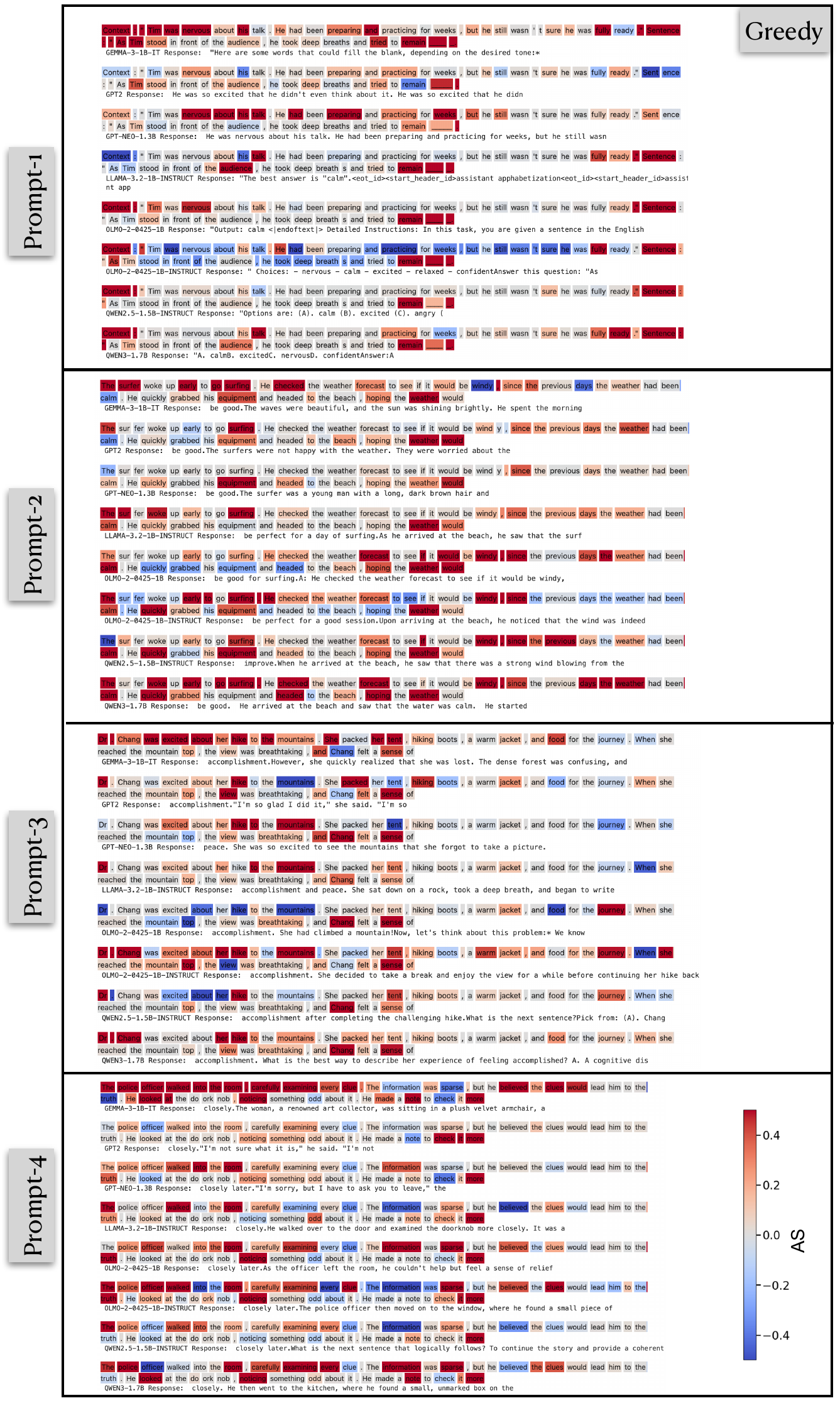}
    \caption{Attribution Scores for prompts given the response for the $8$ LLMs and $4$ prompts with greedy decoding. 
    }
    \label{fig:ASg1}
\end{figure}

\begin{figure}[h]
    \centering
    \includegraphics[width=\linewidth]{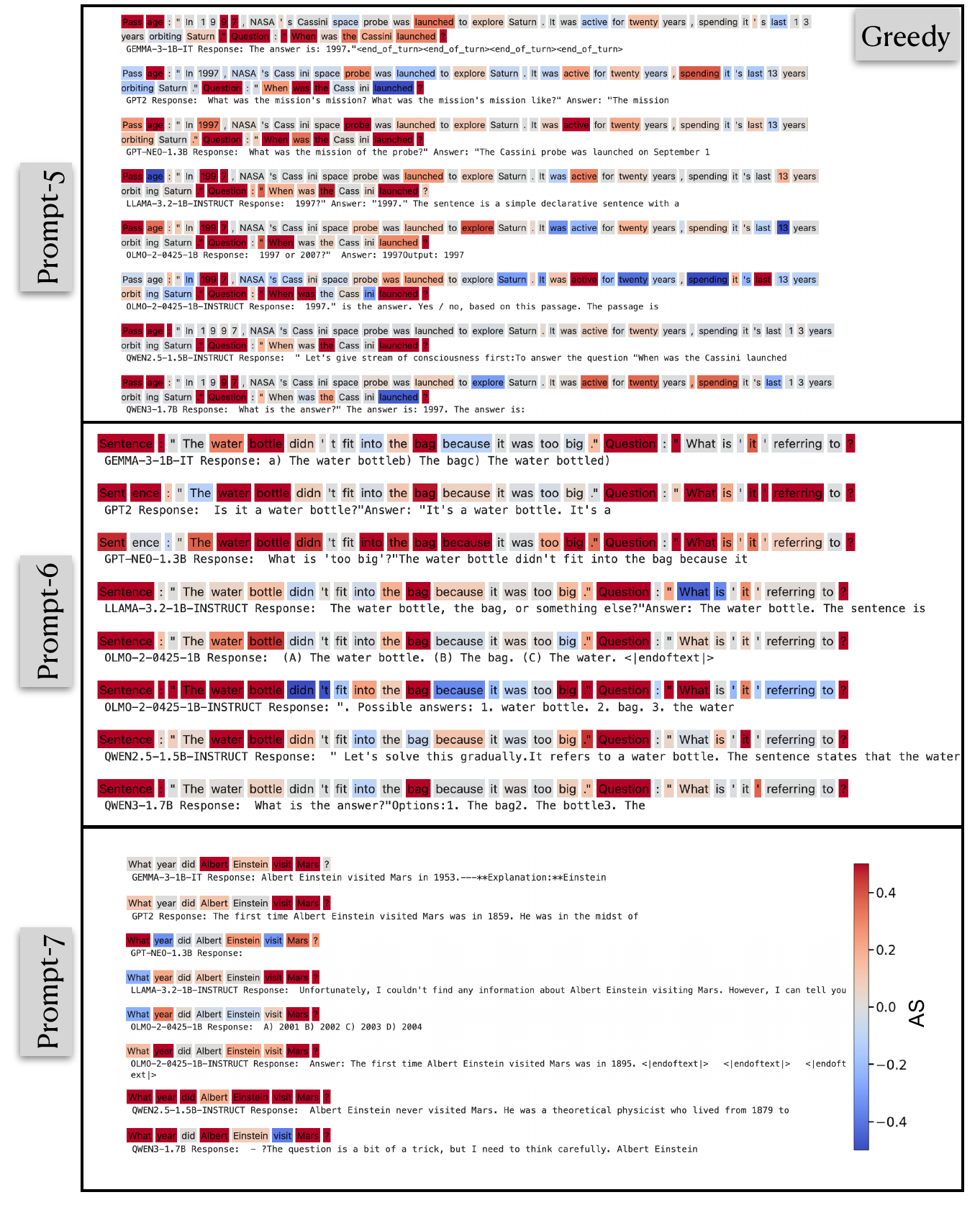}
    \caption{Attribution Scores for prompts given the response for the $8$ LLMs and $3$ prompts with greedy decoding. 
    }
    \label{fig:ASg2}
\end{figure}

\begin{figure}[h]
    \centering
    \includegraphics[height=1.4\linewidth]{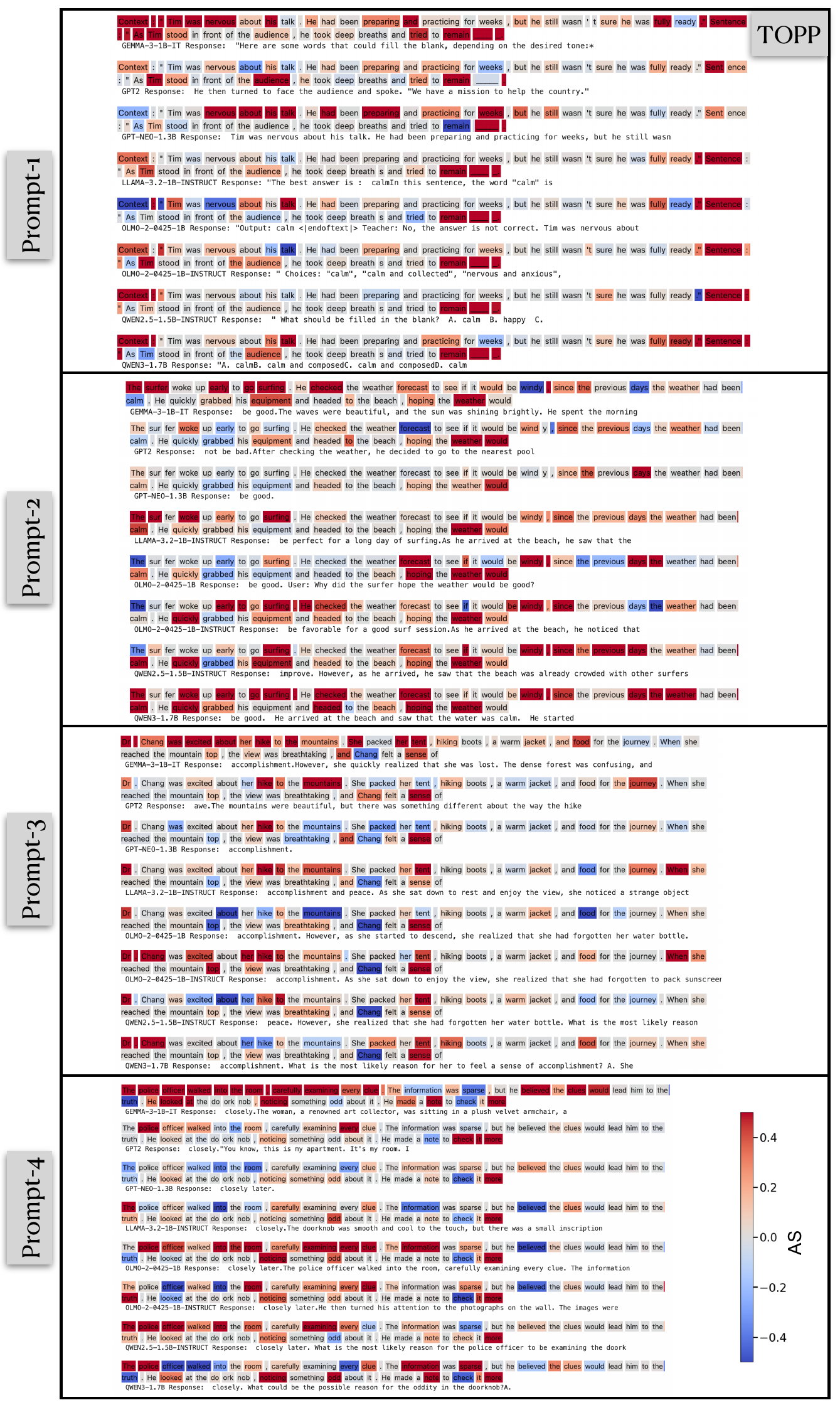}
    \caption{Attribution Scores for prompts given the response for the $8$ LLMs and $4$ prompts  with top-p decoding. 
    }
    \label{fig:ASt1}
\end{figure}

\begin{figure}[h]
    \centering
    \includegraphics[width=\linewidth]{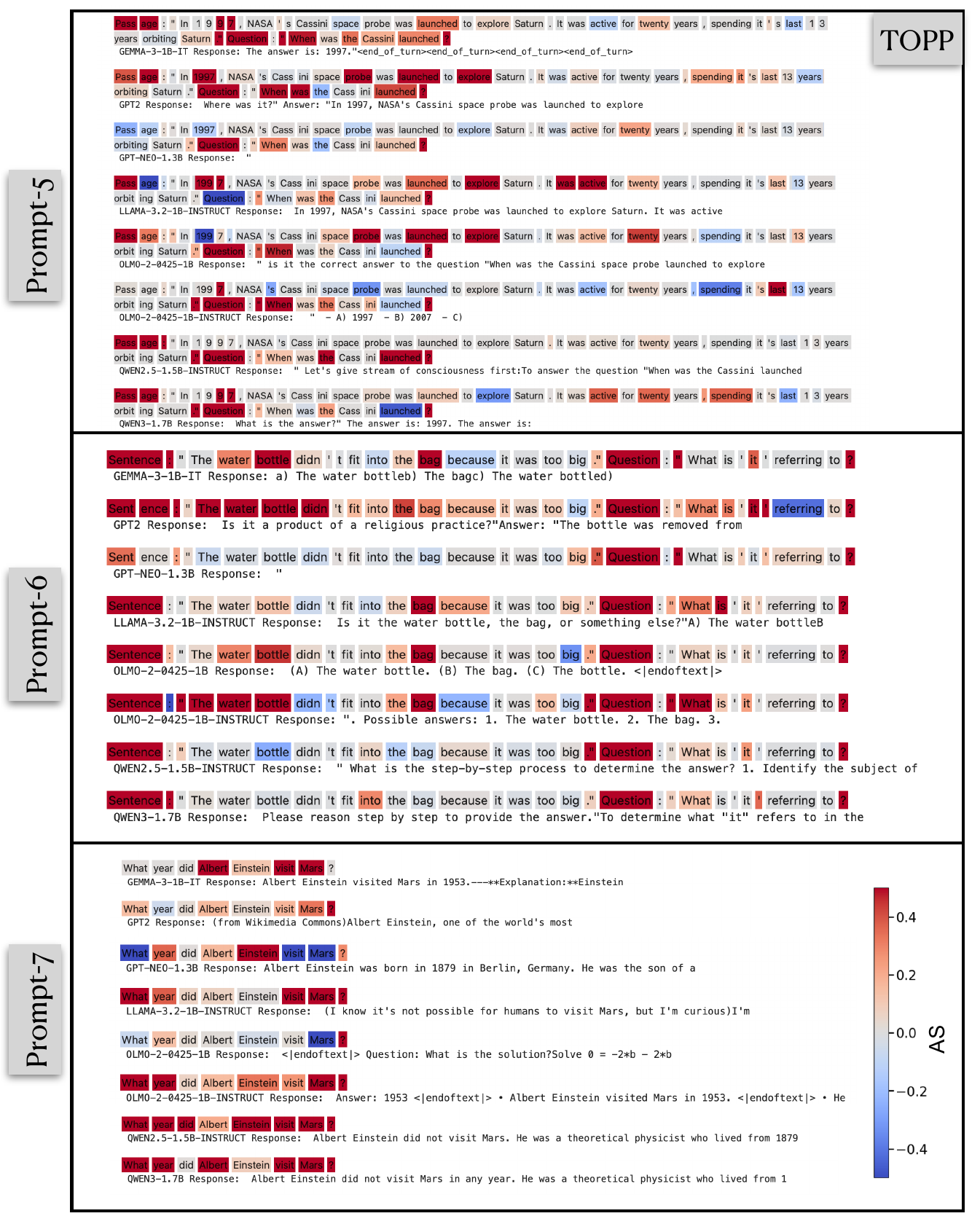}
    \caption{Attribution Scores for prompts given the response for the $8$ LLMs and $3$ prompts with top-p decoding. 
    }
    \label{fig:ASt2}
\end{figure}

\end{document}